\documentclass[final]{cvpr}
\usepackage{times}
\usepackage{epsfig}
\usepackage{graphicx}
\usepackage{amsmath}
\usepackage{amssymb}

\usepackage[dvipsnames]{xcolor}

\usepackage{caption}
\usepackage{subcaption}

\usepackage[pagebackref=true,breaklinks=true,colorlinks,bookmarks=false]{hyperref}

\def\cvprPaperID{10524} 
\def\confYear{CVPR 2021}

\begin{document}

\title{PAUL: Procrustean Autoencoder for Unsupervised Lifting}
\author{Chaoyang Wang$^1$~~~~~~Simon Lucey$^{1,2}$\\
$^1$Carnegie Mellon University   $^2$University of Adelaide\\
{\tt\small \{chaoyanw, slucey\}@cs.cmu.edu}}


\maketitle

\def\cI{\mathcal{I}}
\def\cT{\mathcal{T}}
\def\p{\mathbf{p}}
\def\t{\mathbf{t}}
\def\R{\mathbf{R}}
\def\D{\mathbf{D}}
\def\A{\mathbf{A}}
\def\B{\mathbf{B}}
\def\I{\mathbf{I}}
\def\supi{{(i)}}
\def\bd{\mathbf{d}}
\def\bz{\mathbf{z}}
\def\bw{\mathbf{w}}
\def\M{\mathbf{M}}
\def\S{\mathbf{S}}
\def\W{\mathbf{W}}
\def\w{\mathbf{w}}
\def\x{\mathbf{x}}
\def\t{\mathbf{t}}
\def\cW{\mathcal{W}}
\def\cL{\mathcal{L}}
\def\bvarphi{\boldsymbol{\varphi}}
\def\btheta{\boldsymbol{\theta}}
\def\blambda{\boldsymbol{\lambda}}
\def\bPsi{\mathbf{\Psi}}
\def\bpsi{\boldsymbol{\psi}}

\def\s{\mathbf{s}}
\def\Real{\mathbb{R}}
\def\so{\mathfrak{so}}
\def\SO{\mathbb{SO}}
\def\1{\mathbf{1}}
\def\xy{\text{xy}}
\def\Z{\mathbf{Z}}
\def\d{\mathbf{d}}
\def\C{\mathcal{C}}
\def\tW{\widetilde{\W}}
\def\tD{\widetilde{\D}}
\def\td{\widetilde{\d}}
\def\tbPsi{\widetilde{\bPsi}}
\def\tP{\widetilde{P}}

\def\idx{{(i)}}

\newcommand{\SfM}{S\textit{f}M\xspace}
\newcommand{\SfC}{S\textit{f}C\xspace}
\def\SfMpp{S\textit{f}M++\xspace}

\newcommand{\centered}[1]{\begin{tabular}{l} #1 \end{tabular}}

\begin{abstract}
Recent success in casting Non-rigid Structure from Motion (NR\SfM) as an unsupervised deep learning problem has raised fundamental questions about what novelty in NR\SfM prior could the deep learning offer. In this paper we advocate for a 3D deep auto-encoder framework to be used explicitly as the NRSfM prior. The framework is unique as: (i) it learns the 3D auto-encoder weights solely from 2D projected measurements, and (ii) it is Procrustean in that it jointly resolves the unknown rigid pose for each shape instance. We refer to this architecture as a Procustean Autoencoder for Unsupervised Lifting (PAUL), and demonstrate state-of-the-art performance across a number of benchmarks in comparison to recent innovations such as Deep NRSfM~\cite{ck19} and C3PDO~\cite{c3dpo}.  
\end{abstract}

\section{Introduction}
Inferring non-rigid 3D structure from multiple unsynchronized 2D imaged observations is an ill-posed problem. Non-Rigid Structure from Motion (NR\SfM) methods approach the problem by introducing additional priors -- of particular note in this regards are low rank~\cite{dai2014simple,bregler2000recovering,akhter2009defense} and union of subspaces~\cite{kumar2016multi,zhu2014complex} methods. 

Recently, NRSfM has seen improvement in performance by recasting the problem as an unsupervised deep learning problem~\cite{c3dpo,cha2019unsupervised,park2020procrustean}. These 2D-3D \emph{lifting networks} have inherent advantages over classical NR\SfM as: (i) they are more easily scalable to larger datasets, and (ii) they allow fast feed-forward prediction once trained. These improvement, however, can largely be attributed to the end-to-end reframing of the learning problem rather than any fundamental shift in the prior/constraints being enforced within the NRSfM solution. For example, both Cha~\etal~\cite{cha2019unsupervised} and Park~\etal~\cite{park2020procrustean} impose a classical low rank constraint on the recovered 3D shape. It is also well understood~\cite{dai2014simple,ck19,zhu2014complex,kumar2016multi} that such low rank priors have poor performance when applied to more complex 3D shape variations. 

The NRSfM field has started to explore new non-rigid shape priors inspired by recent advances in deep learning. Kong \& Lucey~\cite{ck19} proposed the use of hierarchical sparsity to have a more expressive shape model while ensuring the inversion problem remains well conditioned. Although achieving significant progress in several benchmarks, the approach is limited by the somewhat adhoc approximations it employs so as to make the entire NRSfM solution realizable as a feed-forward lifting network. Such approximations hamper the interpretability of the method as the final network is a substantial departure from the actually proposed objective. We further argue that this departure from the true objective also comes at the cost of the overall effectiveness of the 2D-3D lifting solution.

In this paper we propose a prior that 3D shapes aligned to a common reference frame are compressible with an undercomplete auto-encoder. This is advantageous over previous linear methods, because the deeper auto-encoder is naturally capable of compressing more complicated non-rigid 3D shapes. What makes learning such an auto-encoder challenging is: (i) it observes only 2D projected measurements of the non-rigid shape; (ii) it must automatically resolve the unknown rigid transformation to align each projected shape instance. We refer to our solution as a \emph{Procustean Autoencoder for Unsupervised Lifting} (PAUL). PAUL is considered unsupervised as it has to handle unknown depth, shape pose, and occlusions. Unlike Deep NRSfM~\cite{ck19}, the optimization process of PAUL does not have to be realizable as a feed-forward network -- allowing for a solution that stays tightly coupled to the proposed mathematical objective.

 We also explore other alternative deep shape priors such as: decoder only and decoder + low-rank. A somewhat similar approach is recently explored by Sidhu~\etal~\cite{Sidhu2020} for dense NR\SfM. Our empirical results demonstrate the fundamental importance of the auto-encoder architecture for 2D-3D lifting.   

\noindent \textbf{Contributions:} We make the following contributions:
\begin{itemize}
    \item We present an optimization objective for joint learning the 2D-3D lifting network and the Procrustean auto-encoder solely from 2D projected measurements.
    \item A naive implementation of PAUL through gradient descent would result in poor local minima, so instead we advocate for a bilevel optimization, whose lower level problem can be efficiently solved by orthographic-N-point (OnP) algorithms.
    \item Our method achieves  state-of-the-art performance across multiple benchmarks, and is empirically shown to be robust against the choice of hyper-parameters such as the dimension of the latent code space.  
\end{itemize}

\begin{figure}[t]
\small
    \centering
    \includegraphics[width=0.9\linewidth]{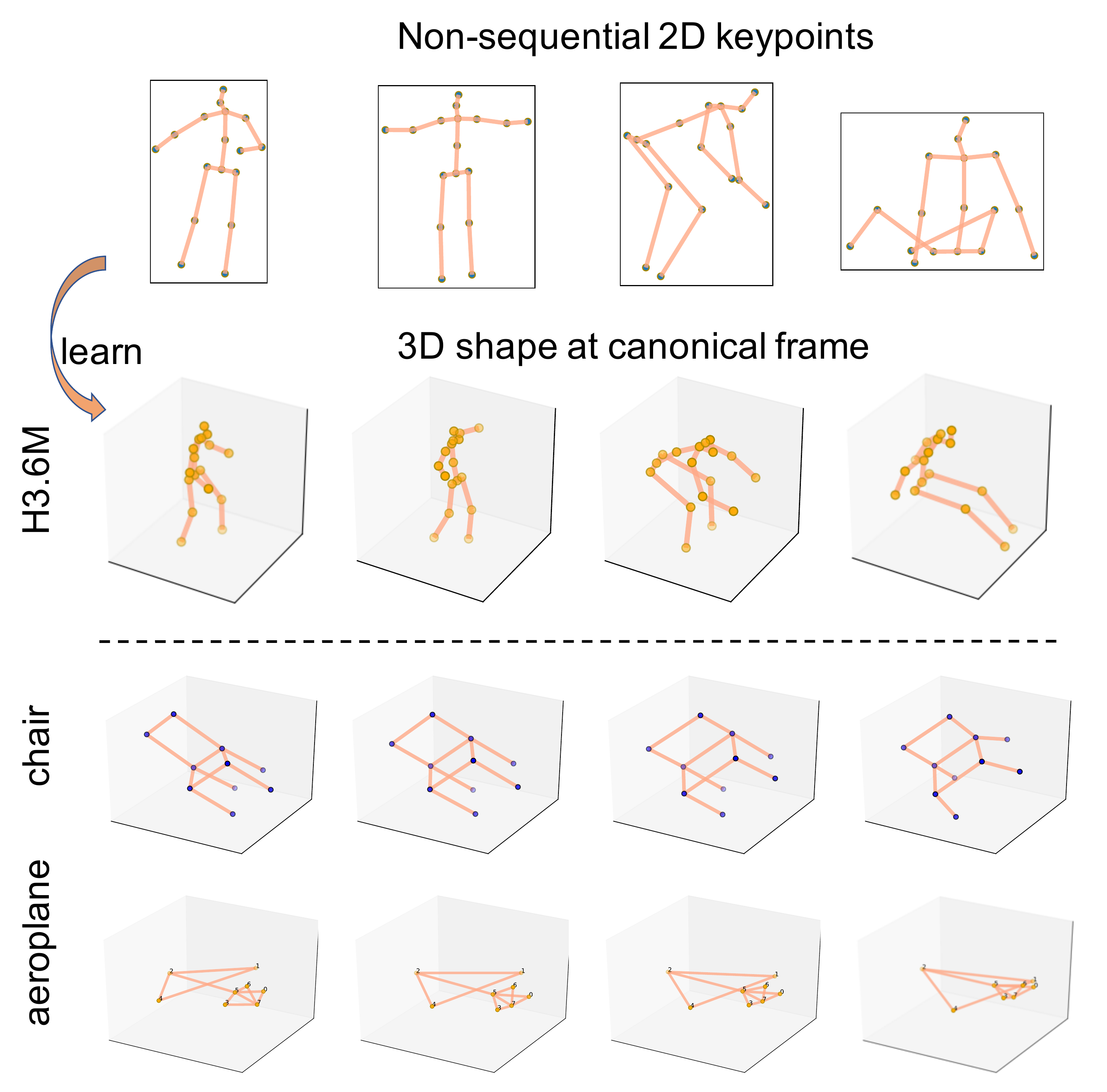}
    \caption{PAUL learns to reconstruct 3D shapes aligned to a canonical frame, using 2D keypoint annotations only. Bottom rows show 3D shapes interpolated from the learned latent space of the Procrustean auto-encoder.}
    \label{fig:teaser}
\end{figure}





\begin{figure*}
\small
    \centering
    \includegraphics[width=\linewidth]{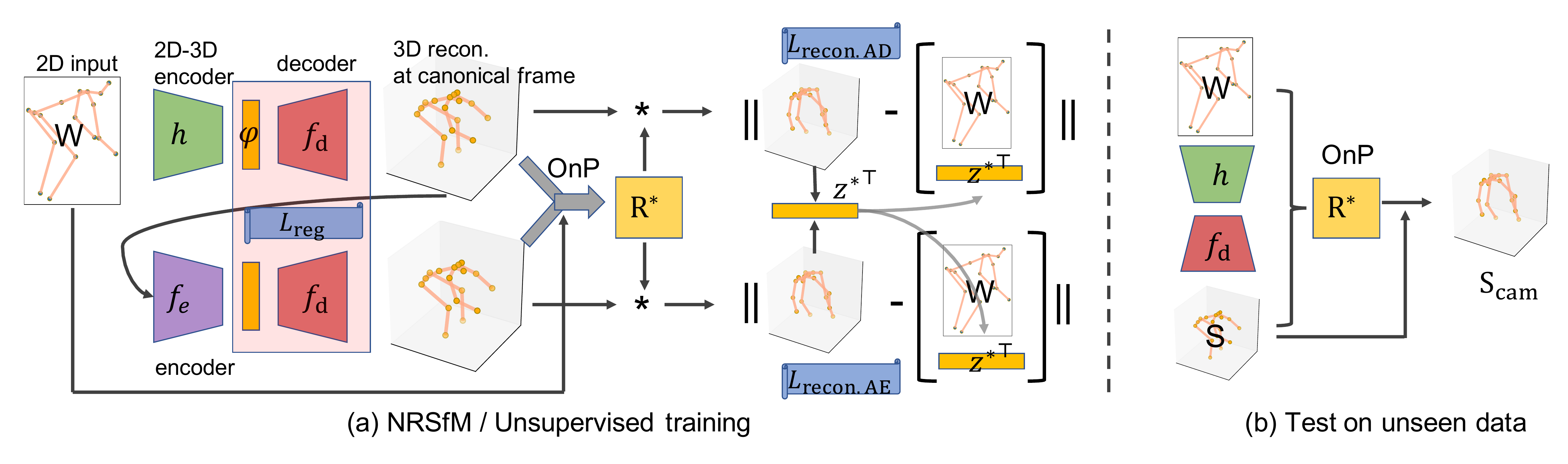}
    \caption{(a) In training, PAUL jointly optimizes the depth value $\mathbf{z}$, camera rotation $\R$ together with the network weights. It is realized through a bilevel optimization strategy, which analytically computes $\R^*$ and $\mathbf{z}^*$ as the solution to an OnP problem. The learning objective is formulated as a combination of reconstruction loss for the decoder-only stream (top row) and the auto-encoder (bottom row) together with regularizer $\mathcal{L}_\text{reg}$ applied on the code $\bvarphi$ and decoder's network weights; (b) in testing, only 2D-3D encoder $h$ and decoder $f_\text{d}$ are used. Camera rotation is directly estimated by OnP. }
    \label{fig:pipeline}
\end{figure*}

\section{Related Work}
\vspace{6px}
\noindent \textbf{Non-rigid structure from motion. }
NR\SfM concerns the problem of recovering 3D shapes from 2D point correspondences from multiple images, \emph{without} the assumption of the 3D shape being rigid. It is ill-posed by nature, and additional priors are necessary to guarantee the uniqueness of the solution. We focus our discussion on the type of priors imposed on shape/trajectory:\\
(i) \emph{low-rank} was advocated by Bregler \etal~\cite{bregler2000recovering} based on the insight that rigid 3D structure has a fixed rank of three~\cite{tomasi1992shape}. Dai~\etal~\cite{dai2014simple} proved that the low-lank assumption is standalone sufficient to solve NR\SfM. It is also applied temporally~\cite{fragkiadaki2014grouping,akhter2009defense} to constraint 3D trajectories. Kumar~\cite{kumar2020non} recently revisited Dai's approach~\cite{dai2014simple} and showed that by properly utilizing the assumptions that deformation is smooth over frames, it is able to obtain competitive accuracy on benchmarks. However, since the rank is strictly limited by the minimum of the number of points and frames~\cite{dai2014simple}, it becomes infeasible to solve large-scale problems with complex shape variations when the number of points is substantially smaller than the number of frames~\cite{ck19}.\\
 (ii) \emph{union-of-subspaces}
 is inspired by the intuition that complex non-rigid deformations could be clustered into a sequence of simple motions~\cite{zhu2014complex}. It was extended to spatial-temporal domain~\cite{kumar2016multi} and structure from category~\cite{Agudo_2018_CVPR}.
 The main limitation of using union-of-subspaces is how to effectively cluster deforming shapes from 2D measurements, and how to compute affinity matrix when the number of frames is huge.\\
(iii) \emph{sparsity}~\cite{kong2016structure,kong2016prior,zhou2016sparseness}, is a more generic prior compared to union-of-subspaces. However, due to the sheer number of possible subspaces to choose, it is sensitive to noise.\\
(iv) \emph{Procrustean normal distribution}~\cite{lee2013procrustean} assumes that the 3D structure follows a normal distribution if aligned to a common reference frame. It allows reconstruction without specifying ranks which are typically required by other methods. It was extended temporally as a Procrustean Markov process~\cite{Lee_2014_CVPR}. Limited by assuming normal distribution, it is less favorable to model deformation which is not Gaussian.\\
\noindent \textbf{Unsupervised 2D-3D lifting.} 
NR\SfM can be recast for unsupervised learning 2D-3D lifting. Cha~\etal~\cite{cha2019unsupervised} use low-rank loss as a learning objective to constraint the shape output of the 2D-3D lifting network. Park~\etal~\cite{park2020procrustean} further modifies Cha's approach by replacing the camera estimation network with an analytic least square solution which aligns 3D structures to the mean shape of a sequence. Due to the inefficiency of low-rank to model complex shape variations, these methods are restricted to datasets with simpler shape variations, or requires temporal order so as to avoid directly handling global shape variations.\\
\indent Instead of using classical NR\SfM priors, recent works explore the use of deeper constraints. Generative Adversarial Networks (GANs)~\cite{goodfellow2014generative} are used to enforce realism of 2D reprojections across novel viewpoints~\cite{chen_drover,repnet,drover,posegan}. These methods are only applicable for large datasets due to the requirement of learning GANs. It is also unclear how to directly learn GANs with training set existing missing data.\\
\indent Novotny~\etal~\cite{c3dpo} instead enforces self-consistency on the predicted canonicalization of the randomly perturbed 3D shapes. Kong \& Lucey~\cite{ck19} proposed the use of hierarchical sparsity as constraint, and approximate the optimization procedure of hierarchical sparse coding as a feed-forward lifting network. It was recently extended by Wang~\etal~\cite{wang2020deep} to handle missing data and perspective projection. These approaches use complicated network architecture to enforce constraints as well as estimating camera motion, while our method uses simpler constraint formulation, and realized with efficient solution.

\section{Preliminary}
\vspace{6px}
\noindent \textbf{Problem setup.} We are interested in the atemporal setup for unsupervised 2D-3D lifting, which is a general setup that not only works with single deforming objects, but also multiple objects from the same object category. Specifically, given a non-sequential dataset consist of $N$ frames of 2D keypoint locations $\{\W^{(1)},\dots, \W^{(N)}\}$, where each $\W\in\mathbb{R}^{2\times P}$ represents 2D location for $P$ keypoints, and visibility masks represented as diagonal binary matrices $\{\M^{(1)},\dots, \M^{(N)}\}$, we want to (i) recover the 3D locations for every keypoints in the dataset, and (ii) train a 2D-3D lifting network capable of making single frame prediction for unseen data.

\noindent \textbf{Weak perspective camera model.}
We assume weak perspective projections, \ie for a 3D structure $\S$ defined at a canonical frame, its 2D projection is approximated as:
\begin{equation}
    \W \approx s \R_{xy}\S + \t_{xy}
\label{eq:weakpersp_proj}
\end{equation}
where $\R_{xy}\in\mathbb{R}^{2\times3}$, $\t_{xy}\in\mathbb{R}^2$ are the x-y component of a rigid transformation, and $s>0$ is the scaling factor inversely proportional to the object depth if the true camera model is pin-hole. If all 2D points are visible and centered, $\t_{xy}$ could be omitted by assuming the origin of the canonical frame is at the center of the object.  Due to the bilinear form of \eqref{eq:weakpersp_proj}, $s$ is ambiguous and becomes  up-to-scale recoverable only when $\S$ is assumed to follow certain prior statistics. A typical treatment to handle scale is to approximate with orthogonal projection by normalizing the scale of $\W$, setting $s=1$ and leaving $\S$ to be scaled reconstruction. 

\noindent \textbf{Regularized auto-encoder (RAE) for $\S$.} We assume that the 3D shapes, if aligned to a canonical frame, are compressible by an undercomplete auto-encoder with a low-dimensional bottleneck, \ie 
\begin{equation}
    \S \approx f_\text{d} \circ f_\text{e}(\S),
    \label{eq:ae}
\end{equation}
where $f_\text{e}$ is the encoder which maps $\S$ to a $K$-dimensional latent code $\bvarphi\in\mathbb{R}^K$ , $f_\text{d}$ is the decoder function and $\circ$ denotes function composition. 
In this work, we choose deterministic RAE~\cite{Ghoshetal19} instead of variational auto-encoder (VAE)~\cite{kingma2014auto} since RAE is easier to train and still leads to an equally smooth and meaningful latent space. The learning objective for RAE is a combination of reconstruction loss and regularizers on the latent codes as well as the decoder's weights,
\begin{equation}
    \mathcal{L}_\text{RAE}(\mathbf{x}; \btheta_\text{d}, \btheta_\text{e}) = \|f_\text{d}\circ f_\text{e} (\mathbf{x}) - \mathbf{x}\|_F + \mathcal{L}_\text{reg}.
\end{equation}
where $\btheta_\text{d}, \btheta_\text{e}$ are network weights for the auto-encode, $\mathbf{x}$ denote data samples, 
and the regularizer  $\mathcal{L}_\text{reg}$ is picked to be $\|\bvarphi\|_2^2$ and weight decay, which was shown to give comparable performance to VAE when generating images and structured objects~\cite{Ghoshetal19}. 

\noindent \textbf{2D-3D lifting network.}
A 2D-3D lifting network is designed to take input from 2D keypoints and visiblity mask, and outputs 3D keypoint locations. We assume the network architecture is decomposed into two parts (i) a 2D-3D encoder $h$ which maps 2D observations to latent code $\bvarphi$, and (ii) a decoder $f_\text{d}$ (reused from the auto-encoder) to generate 3D shapes from $\bvarphi$. Thus this type of 2D-3D lifting network can be expressed as $f_\text{d}\circ h(\W,\M)$, which is a general form for network architectures used in literature~\cite{ck19,martinez2017simple,c3dpo}.  


\section{Learning Procrustean auto-encoder from 2D}
For clarity, in this section we simplify the problem by assuming all points are visible, which allows removing translational component in \eqref{eq:weakpersp_proj}. Description of handling occlusions are given in Sec.~\ref{sec:missing_data}. Fig.~\ref{fig:pipeline} illustrates the proposed approach.

\subsection{Learning objective}
\label{sec:learning_objective}
\noindent \textbf{Procrustean auto-encoder.}
Directly compressing 3D shapes $\S_\text{cam}$ at camera frame is inefficient due to the inclusion of the degrees of freedom from camera motion. 
Therefore, we choose to impose compressibility on $\S$ at canonical frame as shown in \eqref{eq:ae}. However, learning such auto-encoder from 2D observations requires overcoming several obstacles: (i) due to the objects being non-rigid, the definition of canonical frame is statistical and implicitly represented by the unknown rigid transformations to align $\S_\text{cam}$'s; (ii) choosing canonical frame requires knowing the statistics of $\S_\text{cam}$ which we do not have complete information, since only the first two rows of $\S_\text{cam}$ are given as $\W$ representing the x-y coordinates, while the 3rd row $\mathbf{z}$ representing depth values are unknown; (iii) reconstructing $\S_\text{cam}$ in turn requires the estimation of the rigid transformation as well as the statistical model of the shape. To overcome these, we propose a joint optimization scheme:
\begin{equation}
\resizebox{0.89\hsize}{!}{$
    \min\limits_{\substack{\theta_e, \theta_d, \\  \{\mathbf{z}^\idx\}, \{\R^\idx\in SO(3)\}}} \sum\limits_{i=1}^N \mathcal{L}_\text{RAE}({\R^\idx}^\top \begin{bmatrix} \W^\idx \\ {\mathbf{z}^\idx}^\top \end{bmatrix}; ~\btheta_e, \btheta_d),
    $} \;
\label{eq:pa3d_obj}
\end{equation}
where $\theta_e$, $\theta_d$ are network weights for the auto-encoder, \resizebox{0.12\hsize}{!}{$\R^\top \begin{bmatrix} \W\\ \mathbf{z}^\top \end{bmatrix}$} computes $\S$ at the canonical frame. 

However, \eqref{eq:pa3d_obj} is still steps away from being applicable to unsupervised 2D-3D lifting, since it misses the 2D-3D lifting network module in the objective function, and is difficult to optimize due to the inclusion of unknown rotation matrices in the input of the auto-encoder. In the following, we address these by reparameterizing the learning objective and propose an efficient optimization scheme. 

\noindent \textbf{Reparameterization for learning 2D-3D lifting.}
First we introduce an auxiliary variable as the latent code $\bvarphi$, which satisfies 
\begin{equation}
   f_d(\bvarphi) = \R^\top \begin{bmatrix} \W^\top & \mathbf{z} \end{bmatrix}^\top. 
   \label{eq:condition}
\end{equation}
This leads to transforming \eqref{eq:pa3d_obj} to a constrained optimization objective with \eqref{eq:condition} as the constraint and the input to $f_\text{d}\circ f_\text{e}$ replaced by $f_\text{d}(\bvarphi)$,
\begin{equation}
\small
\begin{aligned}
    \min\limits_{\substack{\theta_\text{e}, \theta_\text{d}, \\ \{\R^\idx\in SO(3)\} \\ \{\mathbf{z}^\idx,~\bvarphi^\idx\}}} & \sum\limits_{i=1}^N \|\underbrace{f_\text{d}\circ f_\text{e} \circ f_\text{d}(\bvarphi^\idx) - {\R^\idx}^\top \begin{bmatrix} \W^\idx \\ {\mathbf{z}^\idx}^\top \end{bmatrix} }_{\mathcal{L}_\text{recon. AE}(\bvarphi,\R,\mathbf{z}; \theta_\text{e}, \theta_\text{d})}\|_F + \mathcal{L}_\text{reg}, \\
    & \text{s.t.}~~f_\text{d}(\bvarphi) = \R^\top \begin{bmatrix} \W^\top & \mathbf{z} \end{bmatrix}^\top. 
\end{aligned}
\label{eq:learning_obj}
\end{equation}
Depending on the type of task, $\bvarphi$ could be either treated as free variables to optimize if the task is to reconstruct the `training' set as a NR\SfM problem, or $\bvarphi$ could be the network output of the 2D-3D encoder \ie $\bvarphi = h(\W; \theta_\text{h})$. We then relax the constrained optimization into an unconstrained one, which allows passing gradients to the weights of the 2D-3D encoder $h$,
\begin{equation}
\small
\begin{aligned}
    \min\limits_{\substack{\theta_\text{h}, \theta_\text{e}, \theta_\text{d}, \\ \{\mathbf{z}^\idx\}, \{\R^\idx\in SO(3)\}}} & \sum\limits_{i=1}^N \mathcal{L}_\text{recon. AE}(h(\W^\idx), \R^\idx, \mathbf{z}^\idx)\\ 
    + & ~~\|f_\text{d}\circ h(\W^\idx) - {\R^\idx}^\top \begin{bmatrix} \W^\idx \\ {\mathbf{z}^\idx}^\top \end{bmatrix}\|_F + \mathcal{L}_\text{reg}.
\end{aligned}
\label{eq:learning_obj}
\end{equation}
This loss function could be understood as the combination of the reconstruction losses for both an auto-encoder and an auto-decoder together with the regularizer from RAE, \ie $\mathcal{L}_\text{recon. AE} + \mathcal{L}_\text{recon. AD} + \mathcal{L}_\text{reg}$.

\begin{figure}
\setlength\tabcolsep{0.5pt}
    \begin{tabular}{@{}l@{}@{}c@{}@{}c@{}@{}c@{}}
    & drink & pickup & shark\\
    \centered{\rotatebox{90}{ADL(baseline)}} &
    \centered{\includegraphics[width=0.3\linewidth]{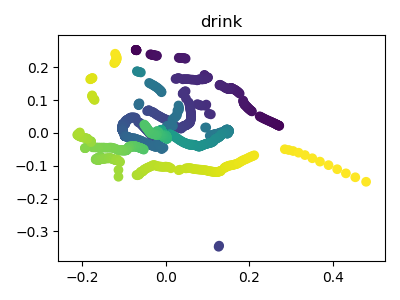}} & 
    \centered{\includegraphics[width=0.3\linewidth]{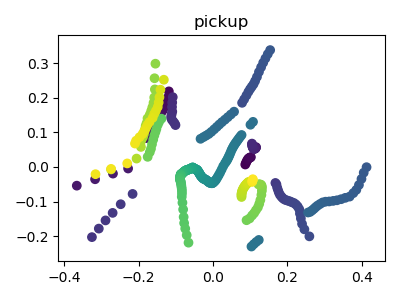}}& 
    \centered{\includegraphics[width=0.3\linewidth]{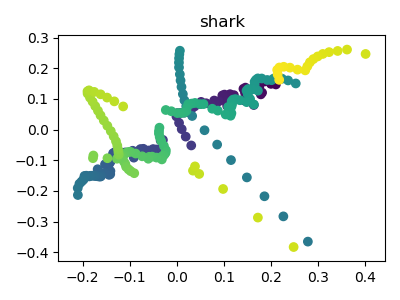}} \\
    \centered{\rotatebox{90}{PAUL}} &
    \centered{\includegraphics[width=0.3\linewidth]{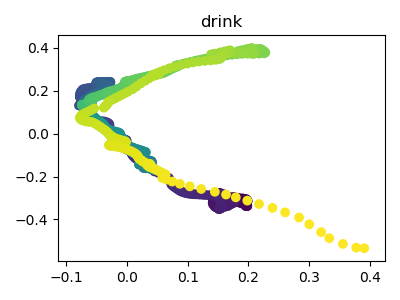}}&
    \centered{\includegraphics[width=0.3\linewidth]{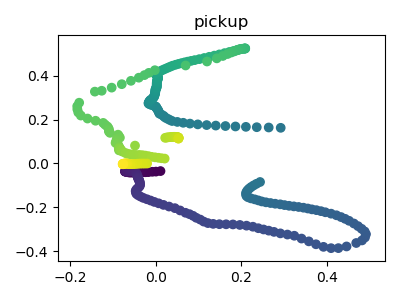}}&
    \centered{\includegraphics[width=0.3\linewidth]{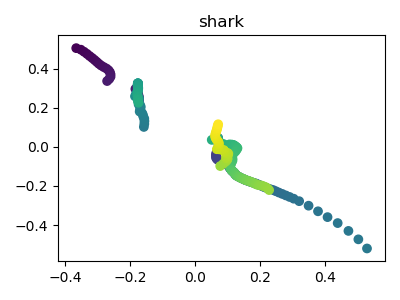}}\\
    \end{tabular}\\
(a) 2D-latent space on \textbf{short} sequences with \textbf{smooth} camera trajectories.\\
\setlength\tabcolsep{4.5pt}
    \begin{tabular}{cc}
         \includegraphics[width=0.45\linewidth]{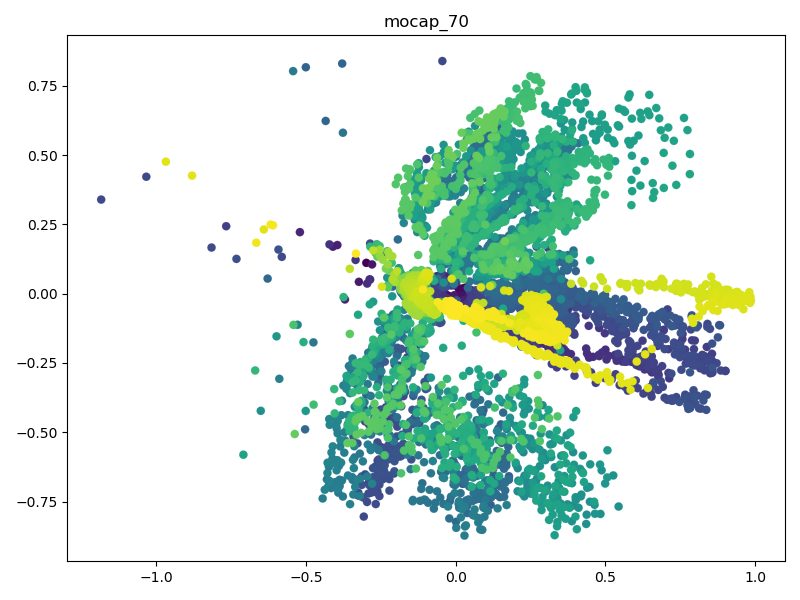} &
         \includegraphics[width=0.45\linewidth]{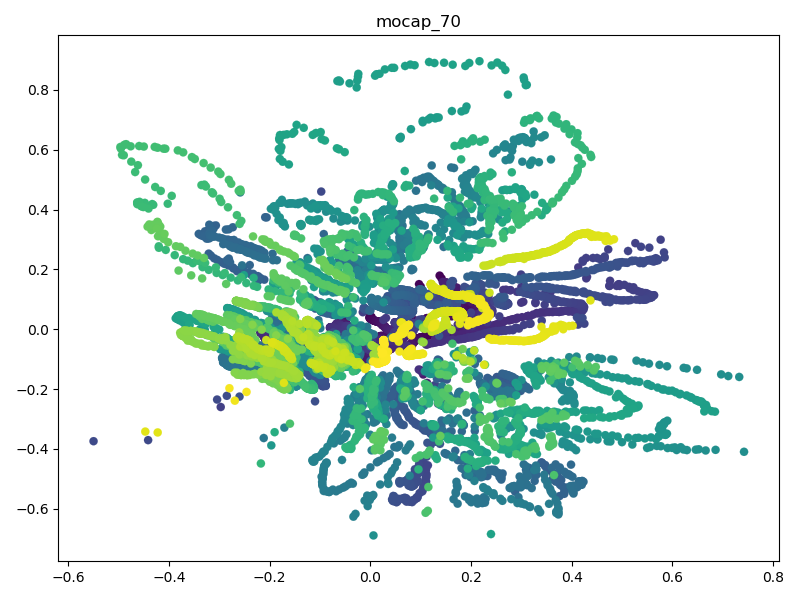} \\
                 ADL (baseline) & PAUL \\
    \end{tabular}
(b) 2D-latent space on \textbf{long} sequence (CMU-MoCap S70) perturbed with \textbf{random} cameras.
    \caption{Visualization of 2D latent space for ADL \& PAUL. Each point represents the 2D latent code recovered for each frame of a sequence. The color of points (from dark blue to bright yellow) indicates the temporal order of points. Ideally, the points should form trajectories in the temporal order.  PAUL gives clearer trajectory-like structures in its latent space, while ADL's recovered codes are either more spread-out or form broken trajectories. }
    \label{fig:latent_space}
\end{figure}

\noindent \textbf{Relation to learning with auto-decoder.} We note that an alternative auto-decoder approach with learning objective $\mathcal{L}_\text{recon. AD} + \mathcal{L}_\text{reg}$ is applicable, the additional $\mathcal{L}_\text{recon. AE}$ in our approach is to enforce the existence of a continuous inverse mapping from 3D shape to latent code. This encourages shapes with small variation to stay close in the latent space, which is helpful to learn a meaningful and smoother latent space. We investigate both approaches in Fig.~\ref{fig:ad_vs_low_rank_vs_paul} and compare the learnt latent space visually in Fig.~\ref{fig:latent_space}.

\subsection{Efficient bilevel optimization}
Directly optimizing \eqref{eq:learning_obj} with gradient descent is inefficient due to (i) the objective is non-convex and it is prone to poor local minima especially with respect to $\R$. One could use an off-the-shelf NR\SfM method to provide initialization for $\R$~\cite{Sidhu2020}. However this would make the solution sensitive to the accuracy of the chosen NR\SfM algorithm.  (ii) when using SGD for large datasets, it is problematic to properly update $\R^\idx$ and $\mathbf{z}^\idx$ if they are left as independent variables. Alternatively, one could introduce additional networks to output $\R$ or $\mathbf{z}$ conditioned on 2D inputs~\cite{repnet,ck19}. We find this unnecessary because it introduces extra complexity to solve the problem but is still subject to the inefficiency of gradient descent. 

For a more efficient optimization strategy, we propose to first rearrange \eqref{eq:learning_obj} to an equivalent bilevel objective:
\begin{equation}
    \min_{\theta_\text{h},\theta_\text{d},\theta_\text{e}} \sum\limits_{i=1}^N \min_{\R^\idx\in SO(3),~\mathbf{z}^\idx} \mathcal{L}^\idx_\text{recon. AE} + \mathcal{L}^\idx_\text{recon. AD} + \mathcal{L}^\idx_\text{reg},
\label{eq:bilevel}
\end{equation}
The benifit of this rearrangement is that the lower level problem, \ie minimizing the reconstruction losses with respect to $\R$ and $\mathbf{z}$ can be viewed as an extension of the orthographic-N-point (OnP) problem~\cite{steger2018algorithms}, which allows the use of efficient solvers~\cite{gower2004procrustes,bojanczyk1999procrustes,mooijaart1990general}. In addition, if an OnP solver refined by geometric loss is able to converge to local minima, it is not required to be differentiable due to the fact that both lower-level and upper-level problems share the same objective function, thus the gradient is zero at local minima~\cite{gould2019deep}. This would lift the restrictions for the type of solvers we could use for the lower-level problem. 

\noindent \textbf{Differentiable fast solver for the lower-level problem.}
On the other hand, we opt to use an algebraic solution which is computationally more light-weight compared to OnP solvers iteratively minimizing the geometric error. 
The compromise of using an approximate (\eg algebraic) solution is that, since it does not not necessarily reach local minima, it is required to be implemented as a differentiable operator, which could be easily accomplished via modern autograd packages. The solution we picked is:
\begin{enumerate}
    \item Find the closed-form least square solution $\Tilde{\R}^*$ for minimizing the reprojection error:
    \begin{equation}
    \resizebox{0.8\hsize}{!}{$
        \min\limits_{\Tilde{\R}} \|\Tilde{\R} (f_\text{d}\circ f_\text{e} \circ f_\text{d})(\bvarphi) - \W \|^2_2 + \|\Tilde{\R} f_\text{d}(\bvarphi) - \W \|_2^2.
        $}\;
    \end{equation}
    
    \item Project $\Tilde{\R}^*$ to become a rotation matrix $\R^*\in SO(3)$ using SVD.
    \item ${\mathbf{z}^*} =  \frac{1}{2}((f_\text{d}\circ f_\text{e} \circ f_\text{d})^\top(\bvarphi){\mathbf{r}_z^*} +  f_\text{d}(\bvarphi)^\top{\mathbf{r}_z^*})$, which is the closed-form least square solution for minimizing:
    \begin{equation}
    \resizebox{0.8\hsize}{!}{$
        \min\limits_{\mathbf{z}} \|(f_\text{d}\circ f_\text{e} \circ f_\text{d})^\top(\bvarphi){\mathbf{r}_z^*} - \mathbf{z} \|^2_2 + \| f_\text{d}(\bvarphi)^\top{\mathbf{r}_z^*} - \mathbf{z}\|_2^2,
        $}\;
    \end{equation}
    where ${\mathbf{r}_z^*}^\top$ denotes the 3rd row of $\R^*$.
\end{enumerate}

\noindent \textbf{End-to-end training.} Finally, with the approximate solution $\R^*$, $\mathbf{z}^*$ for the lower-level problem, the learning objective once again becomes a single level one, which is identical to \eqref{eq:learning_obj} except that $\R$, $\mathbf{z}$ instead of being free variables, they are now replaced by $\R^*$, $\mathbf{z}^*$ which are differentiable functions conditioned on the network weights $\theta_\text{h}, \theta_\text{e}, \theta_\text{d}$. This allows learning these weights end-to-end via gradient descent.

\noindent \textbf{Prediction on unseen data.} To make 3D prediction of a single frame from unseen data, we first use the learned 2D-3D encoder $h$ and the decoder $f_\text{d}$ to compute $\S$ at the canonical frame, and then run OnP algorithm~\cite{steger2018algorithms} to align it to the camera frame.

\subsection{Handling missing data}
\label{sec:missing_data}
If there exists 2D keypoints missing from the observation due to occlusions or out of image, the translational component in \eqref{eq:weakpersp_proj} is no longer removable simply by centering the visible 2D points. To avoid reintroducing $\mathbf{t}$ which would complicate derivations, we choose to follow the object centric trick to absorb translation through adaptively normalizing $\S$ according to the visibility mask $\M$~\cite{wang2020deep}. The normalized $\Tilde{\S}$ is computed as:
\begin{equation}
    \Tilde{\S} = \S + \S (\I_P - \M) \mathbf{1}_P\mathbf{1}_P^\top.
\end{equation}
with this, the projection equation remains bilinear, \ie $\Tilde{\W} = \R_\text{xy} \Tilde{\S}$, where $\Tilde{\W}$ denotes the centered $\W$ by the average of visible 2D points. This allows to adapt PAUL to handle missing data with minimal changes. The detailed description is provided in the supp. material.

\begin{figure}[t]
\small
    \centering
    \includegraphics[width=0.6\linewidth]{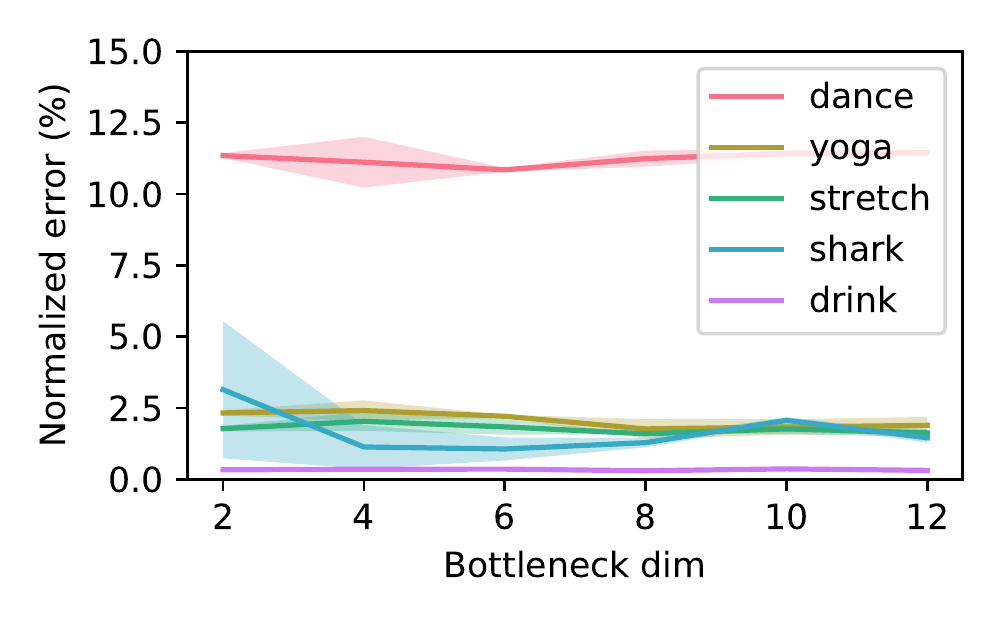}
    \caption{3D reconstruction error with different bottleneck dimensions. For each configuration, PAUL is run 10 times and visualize with average accuracy (solid lines) together with standard deviation (colored area).}
    \label{fig:bottleneck}
\end{figure}

\begin{figure}[t]
\small
    \centering
    \begin{tabular}{ccc}
        \includegraphics[trim={30 30 30 30},clip, width=0.25\linewidth]{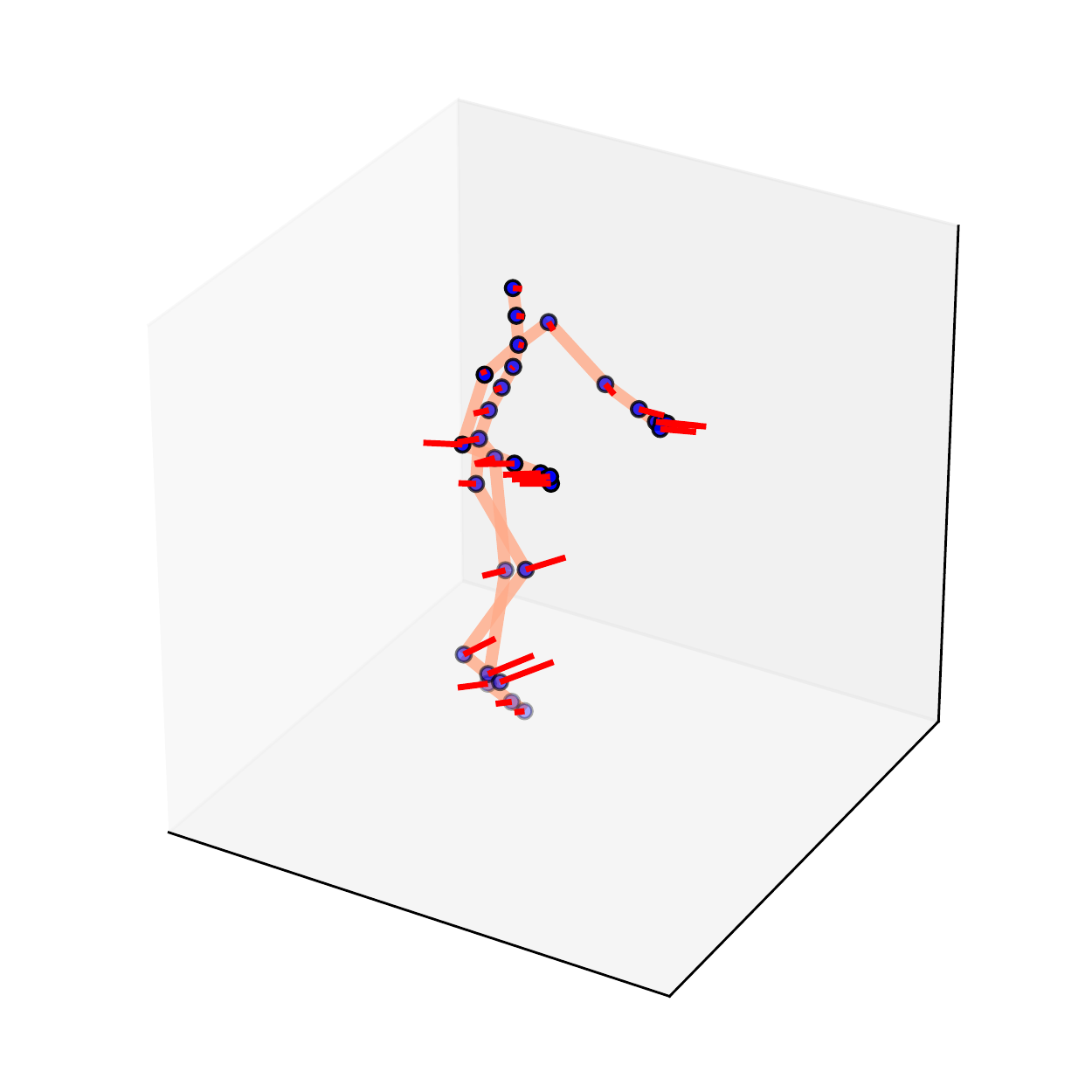} &
        \includegraphics[trim={30 30 30 30},clip, width=0.25\linewidth]{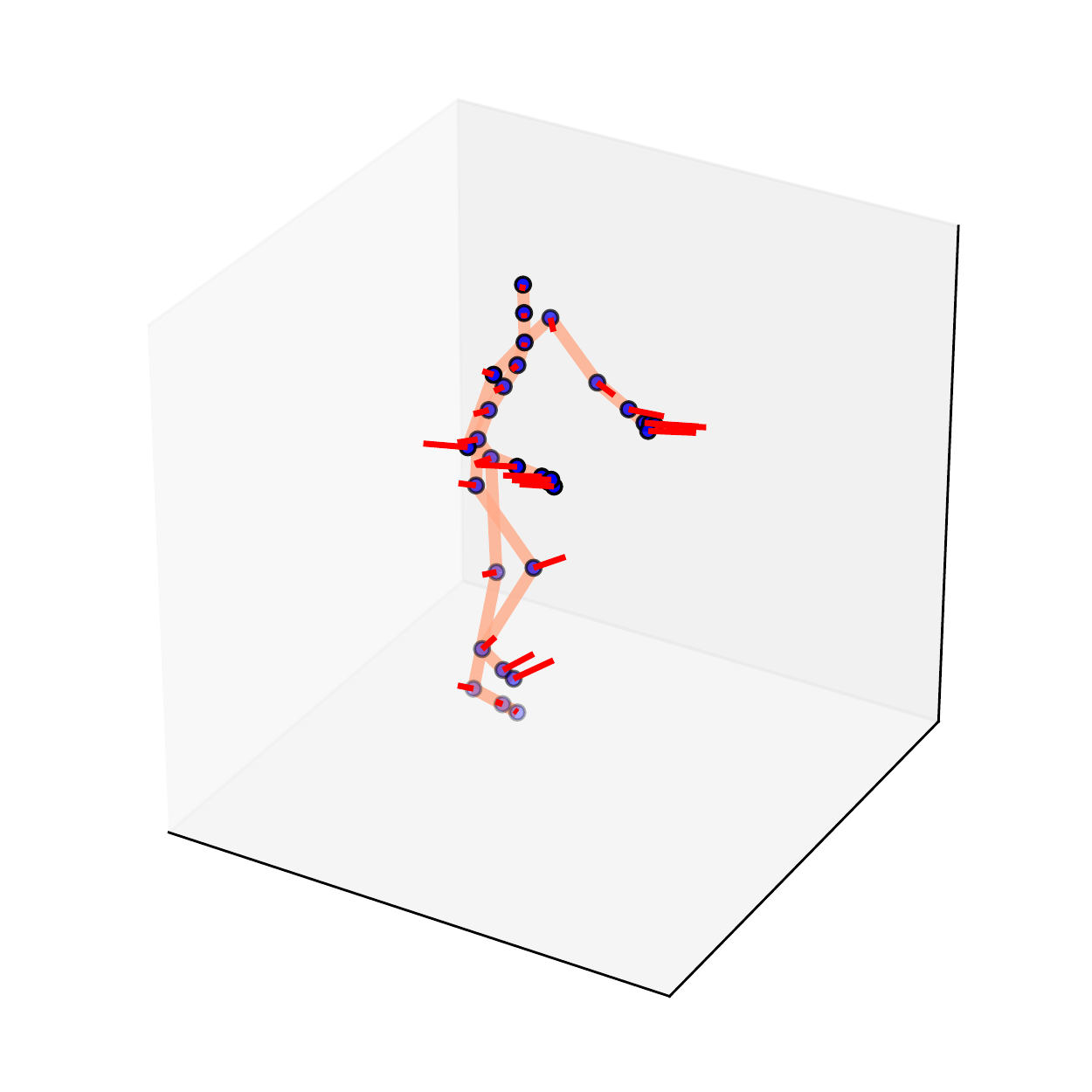} & 
        \includegraphics[trim={30 30 30 30},clip, width=0.25\linewidth]{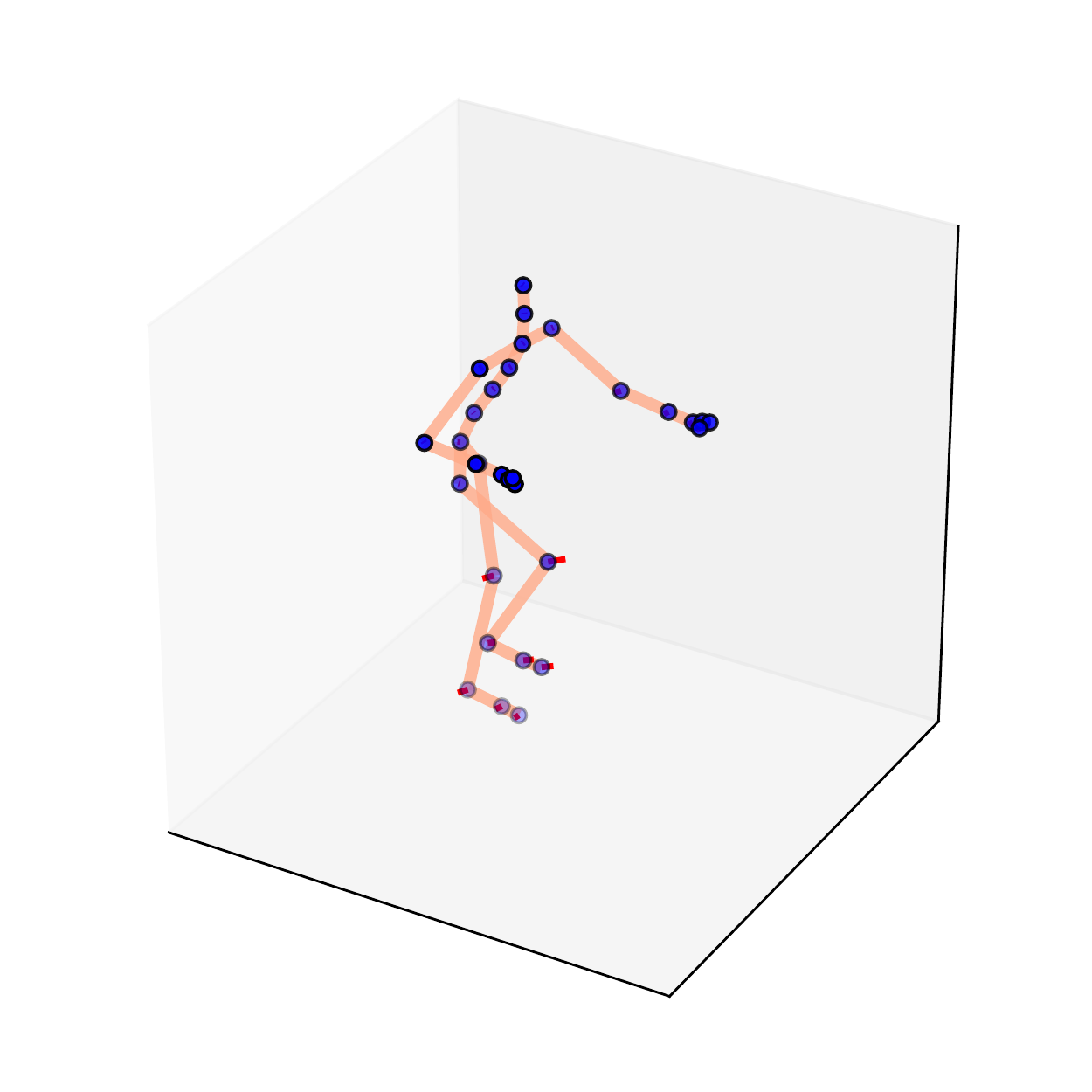}\\
        \includegraphics[trim={30 30 30 30},clip, width=0.25\linewidth]{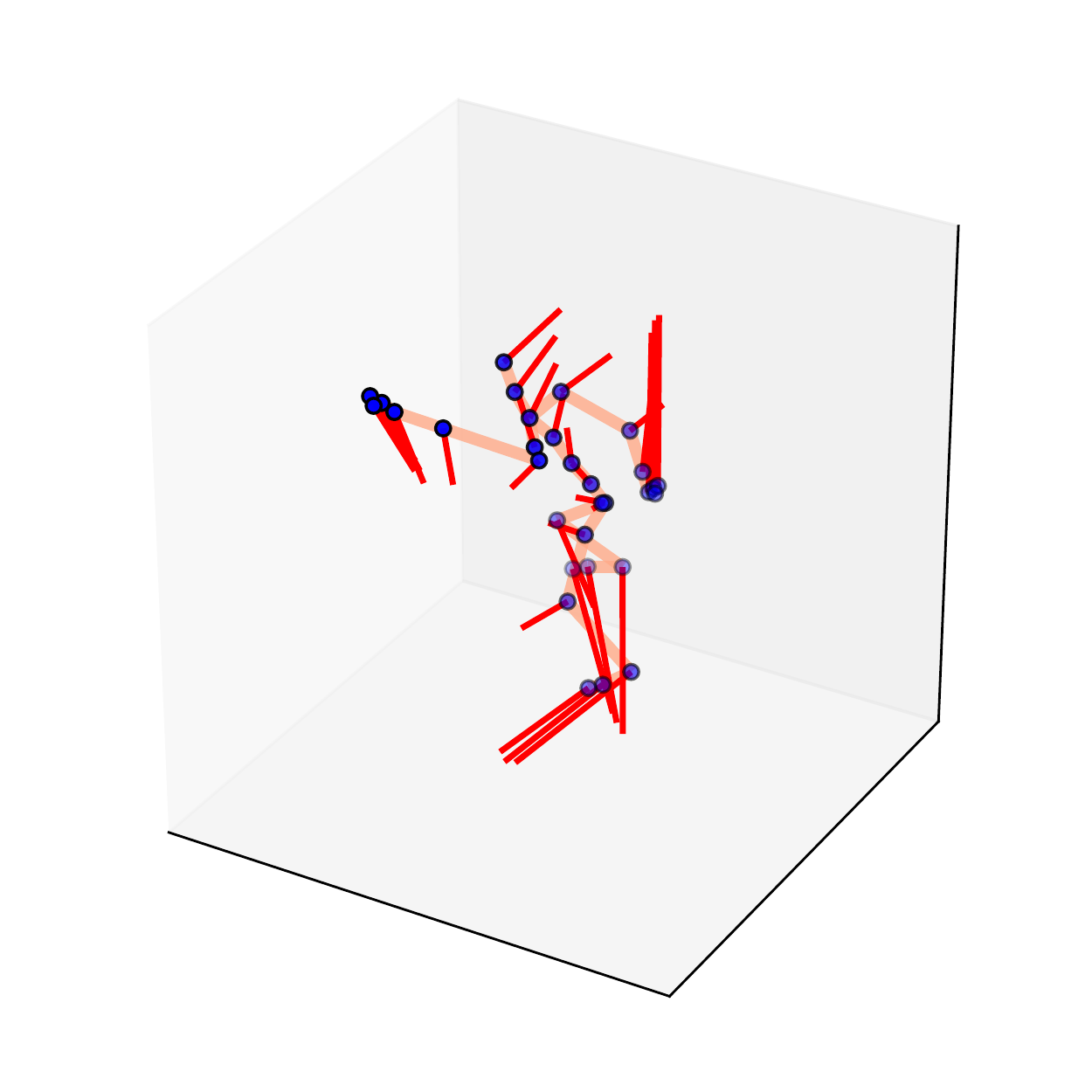} &
        \includegraphics[trim={30 30 30 30},clip, width=0.25\linewidth]{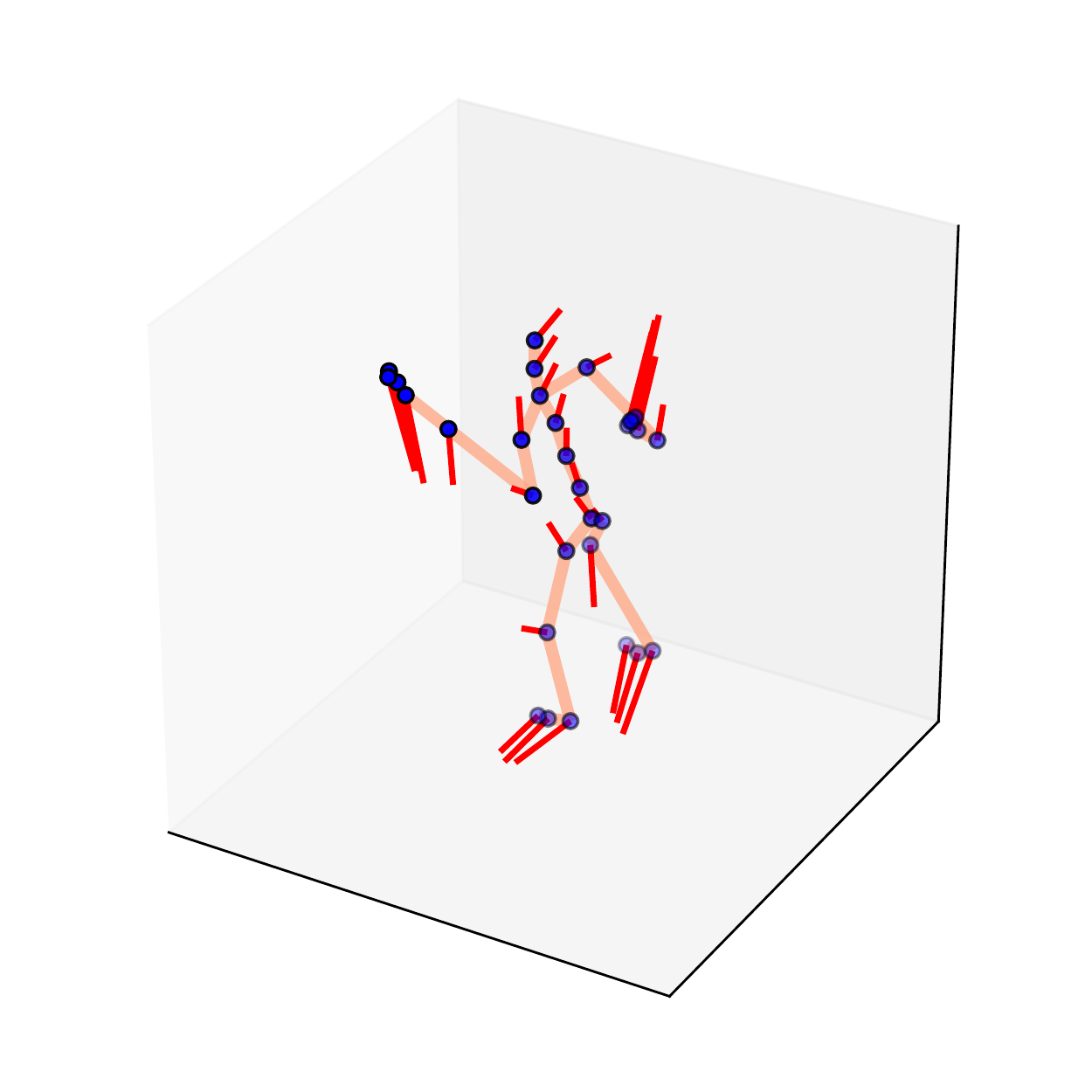} & 
        \includegraphics[trim={30 30 30 30},clip, width=0.25\linewidth]{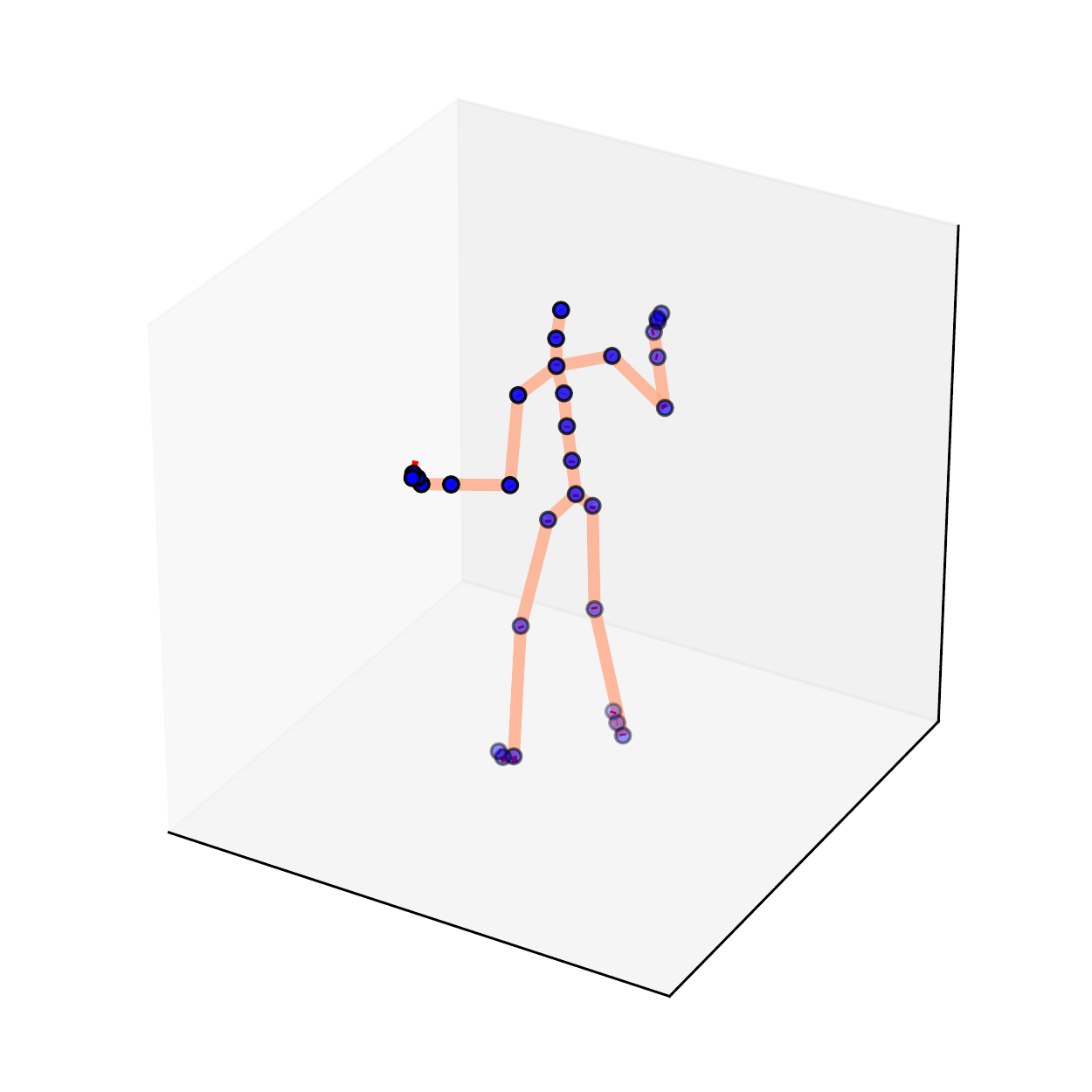}\\
        ADL & ADL + low rank & PAUL(ours) \\
    \end{tabular}
    \includegraphics[width=0.9\linewidth]{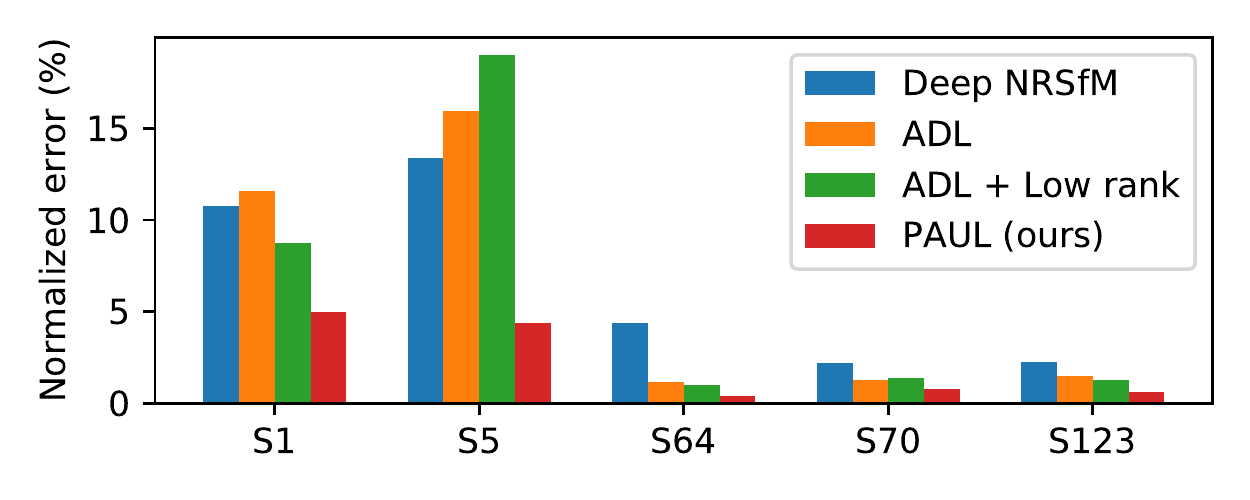}
    \setlength\tabcolsep{0.5pt}
    \caption{Comparison with auto-decoder baseline (\ie ADL), and low rank constraint (ADL + low-rank) on CMU motion capture dataset. PAUL gives significantly more accurate reconstruction compared to ADL and low-rank. \textcolor{red}{Red line} visualizes the difference between reconstructed and groundtruth points.}
    \label{fig:ad_vs_low_rank_vs_paul}
\end{figure}
\section{Experiments}
\subsection{Implementation details}
\noindent\textbf{Network architecture.} Throughout our experiment, we use the same auto-encoder architecture across datasets except the bottleneck dimension. The number of neural units in each layer is decreased exponentially, \ie $\{256, 128, 64, 32, 16\}$. Ideally, if validation set with 3D groundtruth is provided, we could select optimal architecture based on cross validation. However, due to the unsupervised setting, we rather set the hyperparameters heuristically. We pick a smaller bottleneck dimension, \ie 4 for smaller datasets (\eg synthetic NRSfM benchmarks) or datasets with mostly rigid objects (\eg Pascal3D+), and pick a larger dimension, \ie 8 for articulated objects such as human skeleton (H3.6M, CMU motion capture dataset) and meshes (UP3D). The robustness of our method against variations in hyperparameter settings is investigated in Sec.~\ref{sec:nrsfm_exp}. 

For the 2D-3D encoder, we experiment with both fully connected residual network~\cite{martinez2017simple} and convolutional network~\cite{ck19}. The only modification we make to those architecture is the dimension of their output so as to match the picked bottleneck dimension.

\noindent\textbf{Training details.} We keep the same weightings for $\mathcal{L}_\text{reg}$ across all experiments, \ie
\begin{equation}
    \mathcal{L}_\text{reg} = 0.01 \|\bvarphi\|_2^2 + 10^{-4} \|\btheta_\text{d}\|_2^2. 
\end{equation}
We use the Adam optimizer~\cite{kingma2014adam} for training. The optimization parameters are tuned according to specific datasets so as to guarantee convergence. 

\noindent \textbf{Evaluation metric.}
We follow two commonly used evaluation protocols:\\
(i) \emph{MPJPE} evaluates the mean per-joint position error. To account for the inherent ambiguity from weak perspective cameras, we flip the depth values of the reconstruction if it leads to lower error. To account for the ambiguity in the object distance, we either subtract the average depth values or subtract the depth value of a root keypoint. The latter is used only for H3.6M dataset due to the evaluation convention in literature. \\
(ii) \emph{Normalized error} (\textbf{NE}) evalutes the relative error by: $\|\S_\text{pred} - \S_\text{GT}\|_F / \|\S_\text{GT}\|_F$.

\subsection{Baselines}
\noindent \textbf{Auto-decoder lifting (ADL).} As discussed in Sec.~\ref{sec:learning_objective}, an alternative approach for unsupervised lifting is only minimizing the loss $\mathcal{L}_\text{recon. AD} + \mathcal{L}_\text{reg}$, without the term $\mathcal{L}_\text{recon. AE}$ for the auto-encoder. Hence for this baseline, we are only training with respect to the decoder, thus regarded as an auto-decoder approach.\\
\noindent \textbf{ADL + low rank.} In addition, we experiment with adding the low rank constraint as another baseline. Similar to Cha \etal~\cite{cha2019unsupervised} and Park \etal~\cite{park2020procrustean}, we evaluate the nuclear norm of the output of the shape decoder as the approximate low rank loss, \ie $\|\S\|_*$. We empirically pick the weighting for the low rank loss as 0.01.

\begin{table*}[t]
\small
    \centering
    \begin{tabular}{l|c c c c c c | c c c c c}
    \hline
        & \multicolumn{6}{c}{short sequences} & \multicolumn{5}{c}{long sequences (random cam. motion)}\\
        & drink & pickup & yoga & stretch & dance & shark & S1 & S5 & S64 & S70 & S123\\ 
        \#frames & 1102 & 357 & 307 & 370 & 264  & 240 &  45025 & 13773 & 11621 & 10788 & 10788\\
        \hline
        \textcolor{OliveGreen}{CNS}~\cite{lee2016consensus} & 3.04 & 9.18 & 11.15 & 7.97 & \textbf{7.59} & 8.32 & 37.62 & 40.02 & 29.00 & 26.26 & 26.46\\
        \textcolor{RedOrange}{PND}~\cite{lee2013procrustean} & \textbf{0.37} & 3.72 & 1.40 & 1.56 & 14.54 & 1.35 & - & - & - & - & - \\
        \textcolor{RedOrange}{BMM}~\cite{dai2014simple} & 2.66 & 17.31 & 11.50 & 10.34 & 18.64 & 23.11 & 16.45 & 14.07 & 18.13 & 18.91 & 19.32\\
        \textcolor{OliveGreen}{BMM-v2}~\cite{kumar2020non} & 1.19 & 1.98 & \textbf{1.29} & \textbf{1.44} & 10.60 & 5.51 & - & - & - & - & - \\
        \textcolor{RedOrange}{Deep NRSfM}~\cite{ck19} & 17.38 & \textbf{0.53} & 12.54 & 21.63 & 20.95 & 21.83 & 10.74 & 13.40  & 4.38 & 2.17 & 2.23\\
        \hline 
        \textcolor{RedOrange}{PAUL} &  0.47 & 2.03 & 1.71 & 1.62 & 10.22 & \textbf{0.37} & \textbf{4.97} & \textbf{4.38} & \textbf{0.39} & \textbf{0.77} & \textbf{0.59} \\
        \hline
    \end{tabular}
    \caption{Comparison with state-of-the-art NR\SfM methods on both short sequences and long sequences, report with normalized error. Long sequences are sampled from CMU motion capture dataset~\cite{cmumocap} with large random camera motion. Atemporal methods are highlighted by \textcolor{RedOrange}{orange}, methods using temporal information are marked by \textcolor{OliveGreen}{green}. Due to the code for PND and BMM-v2 is unavailable, they are excluded from evaluation on CMU motion capture sequences. }
    \label{tab:nrsfm}
\end{table*}
\begin{table*}[t]
\small
    \centering
    \begin{tabular}{l| c c c c c c c c c c c c | c c}
    \hline
         & aero. & car & tv. & sofa & motor. & dining. & chair & bus & bottle & boat & bicycle & train & Mean & 8 cls.\\
         \hline
    C3DPO & 6.56 & 8.21 & 15.03 & 7.30 & 7.48 & \textbf{3.77} & 3.46 & 20.41 & 7.48 & \textbf{7.58} & 3.47 & 33.70 & 10.4 & 7.58\\
    Deep NR\SfMpp& 7.51 & 9.22 & 17.43 & 9.37 & 6.18 & 12.90 & 3.97 & 18.02 & 2.08 & 9.18 & 4.03 & \textbf{23.67} & 10.3 & 8.90\\
    \hline
    PAUL & \textbf{3.99} & \textbf{7.13} & \textbf{9.88} & \textbf{3.99} & \textbf{3.74} &  5.70 & \textbf{2.19} & \textbf{14.11} & \textbf{1.03} & 8.08 & \textbf{1.74} & 38.78 & \textbf{8.4} & \textbf{5.32}\\
    \hline
    \end{tabular}
    \caption{Per-category normalized error (\%) on Pascal3D+ dataset. Follow the protocol of Agudo~\etal~\cite{Agudo_2018_CVPR}, we further report the average error of 8 object categories which are annotated with $\geq 8$ keypoints.}
    \label{tab:pascal3d}
\end{table*}

\begin{table*}[t!]
    \centering
    \parbox{0.4\linewidth}{
        \resizebox{\linewidth}{!}{
            \begin{tabular}{l|c|c}
            \hline
                     & UP3D 79KP & Pascal3D+\\
                avg occlusion \% & 61.89 & 37.68\\
                \hline
                EM-\SfM~\cite{torresani2008nonrigid} & 0.107 & 131.0\\
                GbNR\SfM~\cite{fragkiadaki2014grouping} & 0.093 & 184.6 \\
                \hline
                 Deep NR\SfM~\cite{ck19} & 0.076 & 51.3\\
                 C3DPO~\cite{c3dpo} & 0.067 & 36.6\\
                 Deep NR\SfMpp~\cite{wang2020deep} & 0.062 & 34.8 \\
                 PAUL & \textbf{0.058} & \textbf{30.9}\\
                 \hline
            \end{tabular}
        }
    \caption{Comparison on datasets with high percentage of missing data. Test accuracy is reported with MPJPE.}
    \label{tab:missing_data}
    }
    \hspace{10px}
    \parbox{0.3\linewidth}{
        \resizebox{\linewidth}{!}{
            \begin{tabular}{l|c|c}
            \hline
            & GT pts. & SH pts.~\cite{c3dpo}\\ 
            \hline
                \textcolor{RedOrange}{Pose-GAN}~\cite{posegan} & 130.9 & 173.2\\
                 \textcolor{RedOrange}{C3DPO}~\cite{c3dpo} & 95.6 & 153.0 \\
                 \textcolor{OliveGreen}{PRN}~\cite{park2020procrustean} & 86.4 & 124.5 \\
                 \hline
                 \textcolor{RedOrange}{PAUL} & 88.3 & 132.5\\
                 \hline
            \end{tabular}
        }
    \caption{MPJPE on H3.6M validation set. \textcolor{RedOrange}{orange} indicates atemporal method and \textcolor{OliveGreen}{green} indicates methods use temporal information. }
    \label{tab:h36m}
    }
    \hspace{10px}
    \parbox{0.2\linewidth}{
        \resizebox{\linewidth}{!}{
            \begin{tabular}{l|c}
            \hline
                & NE (\%)\\
                \hline
                \textcolor{RedOrange}{C3DPO}~\cite{c3dpo} &  35.09\\
                \textcolor{OliveGreen}{PRN}~\cite{park2020procrustean} & 13.77\\
                \textcolor{RedOrange}{PAUL} (ours) & \textbf{12.36}\\
                \hline
                \textcolor{RedOrange}{PAUL} (train set) & 4.30\\
                \hline
            \end{tabular}
        }
    \caption{Test accuracy on SURREAL synthetic sequences. Training error is also reported for PAUL (bottom row).}
    \label{tab:surreal}
    }
\end{table*}

\subsection{NR\SfM experiments}
\label{sec:nrsfm_exp}
In the first set of experiments, we evaluate the proposed method for the NR\SfM task, where we report how well the compared methods are able to reconstruct a dataset. The goal is to evaluate the robustness of the proposed Procrustean auto-encoder shape prior across different shape variations, without being convoluted by the inductive bias from a 2D-3D lifting network, which is not the interest of this work. To achieve this, on short sequences, instead of conditioning $\bvarphi$ with a 2D-3D encoder, we treat $\bvarphi$ as free variable to optimize directly; and on long sequences, we use the same 2D-3D encoder as in Deep NR\SfM~\cite{ck19} to have a fair comparison.

\noindent \textbf{NR\SfM datasets.} We report performance on two types of datasets: (i) short sequences with simple object motions, \eg \emph{drink, pickup, yoga, stretch, dance, shark} which are standard benchmarks used in NR\SfM literature~\cite{akhter2009nonrigid,torresani2008nonrigid}.\\ (ii) long sequences with large articulated motions, \ie CMU motion capture dataset~\cite{cmumocap}. We use the processed data from Kong \& Lucey~\cite{ck19} which is intentionally made more challenging by inserting large random camera motions.

\noindent \textbf{Robustness against bottleneck dimension.} As shown in Fig.~\ref{fig:bottleneck}, we run the methods with varying bottleneck dimension from 2 to 12 on different datasets. To account for the stochastic behavior due to network initialization and gradient descent on small datasets, we run the methods 10 times and visualize with average accuracy (solid lines) together with standard deviation (colored area). PAUL gives stable results once the bottleneck dimension is sufficiently large. This indicates that PAUL is practical for unseen datasets by using an overestimated bottleneck dimension. 

\noindent \textbf{Comparison with ADL and low rank.}
As shown in Fig.~\ref{fig:ad_vs_low_rank_vs_paul}, on sequences from CMU motion capture dataset,
ADL achieves lower error in most sequences when comparing against Deep NR\SfM, indicating it is indeed a strong baseline. Augmenting ADL with low rank constraint is able to further decrease error for several sequences, but the improvement is not consistent across the whole dataset. In comparison, PAUL gives significant error reduction for all the evaluated sequences, which demonstrates the effectiveness of the proposed Procrustean anto-encoder prior.

\noindent \textbf{Comparison with state-of-the-art NR\SfM methods.} Table~\ref{tab:nrsfm} collects results from some of the state-of-the-art NR\SfM methods on the synthetic bechmarks, \eg BMM-v2~\cite{kumar2020non}, CNS~\cite{lee2016consensus} and PND~\cite{lee2013procrustean}. All the well-performing methods utilize temporal information while PAUL does not, but still achieves competitive accuracy on short sequences. On long sequences from CMU motion capture dataset, the accuracy of temporal-based methods \eg CNS deteriorates significantly due to the data perturbed by large random camera motion. Atemporal methods on the other hand gives stable results and PAUL outcompetes all the compared methods by a wide margin.


\subsection{2D-3D lifting on unseen data}
We compare against recent unsupervised 2D-3D lifting methods on the processed datasets by Novotny~\etal~\cite{c3dpo}:\\
\noindent \textbf{Datasets.}
(i) \emph{Synthetic UP-3D} is a large synthetic dataset with dense human keypoints collected from the UP-3D dataset~\cite{up3d}. The 2D keypoints are generated by orthographic projection of the SMPL body shape with the visibility computed from a ray traccer. Similar to C3DPO, we report result for 79 representative vertices of the SMPL on the test set;\\
(ii) ~\emph{Pascal3D+}~\cite{xiang2014beyond} consists of images from 12 object categories with sparse keypoint annotations. The 3D keypoint groundtruth are created by selecting and aligning CAD models. To ensure consistency between 2D keypoint and 3D groundtruth, the orthographic projections of the aligned 3D CAD models are used as 2D keypoint annotations, and the visibility mask are taken from the original 2D annotations. For a fair comparison against C3DPO, we use the same fully-connected residual network as the 2D-3D encoder, and train a single model to account for all 12 object categories.\\
(iii) ~\emph{Human 3.6 Million dataset} (H3.6M)~\cite{h36m} is a large-scale human pose dataset annotated by motion capture systems. Following the commonly used evaluation protocol, the first 5 human subjects (1, 5, 6, 7, 8) are used for training and 2 subjects (9, 11) for testing. The 2D keypoint annotations of H3.6M preserves perspective effect, thus is a realistic dataset for evaluating the practical usage of 2D-3D lifting.

\noindent\textbf{Robustness against occlusion.} Both synthetic UP3D and Pascal3D+ dataset simulate realistic occlusions with high occlusion percentage. We focus our comparison against C3DPO and Deep NR\SfMpp~\cite{wang2020deep} which is a recent update of Deep NR\SfM for better handling missing data and perspective projections. As shown in Table~\ref{tab:missing_data}, PAUL significantly outperforms both of them. To account for the distortion caused by the object scale, we switch the evaluation metric from MPJPE to normalized error in Table~\ref{tab:pascal3d} and report per-class error. PAUL leads with even bigger margin.

\noindent\textbf{Robustness against labeling noise.} 
To work with in-the-wild data, 2D-3D lifting methods are required to be robust against annotation noise, which could be simulated by using 2D keypoints detected by a pretrained keypoint detector. In addition, 2D annotation with perspective effect could also be regarded as noise since it is not modeled by the assumed weak perspective camera model. We evaluate both scenarios on H3.6M dataset (see Table~\ref{tab:h36m}). PAUL outperforms the compared atemporal methods (\ie C3DPO and Deep NR\SfMpp) and is competitive to recently proposed PRN~\cite{park2020procrustean} which requires training data to be sequential.

\subsection{Dense reconstruction}
We follow the comparison in Park~\etal~\cite{park2020procrustean} on the synthetic SURREAL dataset~\cite{varol17_surreal}, which consists of 5k frames with 6890 points for training, and 2,401 frames for testing. Unlike PRN~\cite{park2020procrustean} which subsamples a subset of points when evaluating the low rank shape prior due to the intense computational cost of evaluating nuclear norm, our auto-encoder shape prior is computationally cheaper when dealing with dense inputs, thus we made no modification when applying PAUL to SURREAL. As shown in Table~\ref{tab:surreal}, PAUL achieves lower test error compared to PRN, even though we use no temporal information in training. It is worth to point out that the current bottleneck in achieving better test accuracy is at the generalization ability of the 2D-3D encoder network, not at the proposed unsupervised training framework. As shown in the last row of Table~\ref{tab:surreal}, the reconstruction error on the training set is already much lower than the test error (\ie 4.30\% vs 12.36\%). 



\begin{figure}[t!]
\small
    \centering
    \includegraphics[width=\linewidth]{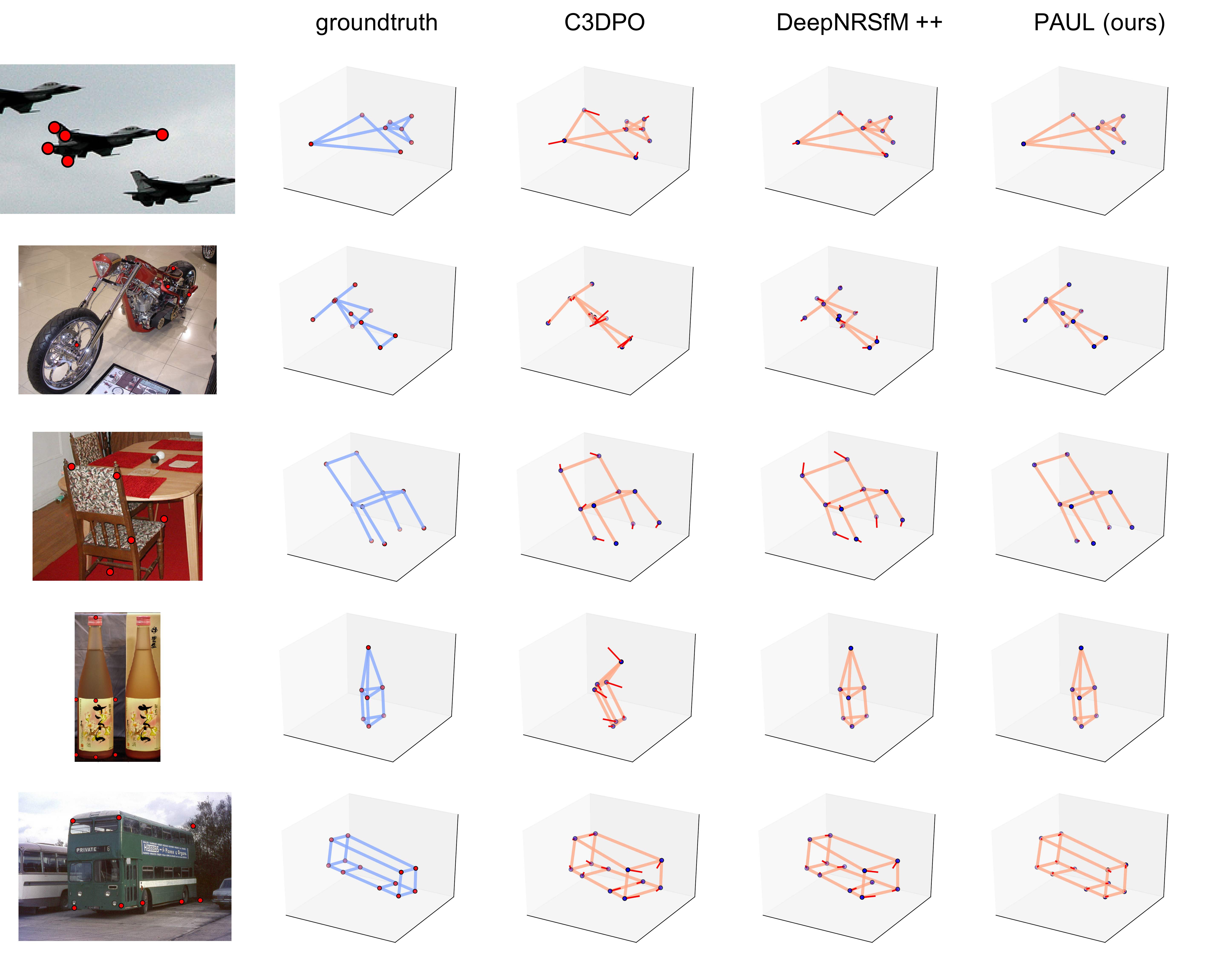}
    \caption{Qualitative comparison on Pascal3D+ dataset. \textcolor{red}{Red lines} visualize the difference between groundtruth points and predicted points. PAUL shows more accurate prediction in the compared samples. }
    \label{fig:pascal}
\end{figure}


\section{Conclusion}
We propose learning a Procrustean auto-encoder for unsupervised 2D-3D lifting capable of learning from no-sequential 2D observations with large shape variations. We demonstrate that having an auto-encoder performs favorably compared to an alternative auto-decoder approach. The proposed method achieves state-of-the-art accuracy across NR\SfM and 2D-3D lifting tasks. For future work, theoretical analysis of the characterization of the solution (\eg uniqueness) may help inspire further development. Interpreting the approach as learning manifold may also help provide guidance such as setting hyperparameters~\cite{camastra2016intrinsic}. Finally, it is straightforward to extend the method to model perspective projection using similar extensions outlined in~\cite{wang2020deep,zheng2013revisiting}.



\noindent\textbf{Acknowledgement} This work was partially supported by the National Science Foundation under Grant No.1925281.


\section*{I. Additional discussion of latent space in PAUL}
PAUL uses the constraint that complex shape variation is compressible into a lower dimensional latent space with an auto-encoder.
The use of latent space in our problem is  different to generative modeling in that: (i) we focus on the compressibility instead of compactness of the representation, \ie we do not require that any linear interpolation between two latent codes still corresponds to a valid 3D shape. (ii) due to the shape coverage from short sequences in NR\SfM is sparse and arbitrary, we do not impose any prior distribution such as Gaussian on the latent space~\cite{zadeh2019variational,kingma2014auto}. However, these do not prevent us from sampling the learned latent space, which can be achieved with an ex-post density estimation step as shown by Ghosh~\etal~\cite{Ghoshetal19}.

\begin{figure*}[h]
\centering
\begin{tabular}{c c c c c c}
    (a) drink & (b) pickup & (c) yoga & (d) stretch & (e) dance  & (f) shark \\
    \includegraphics[width=0.13\linewidth]{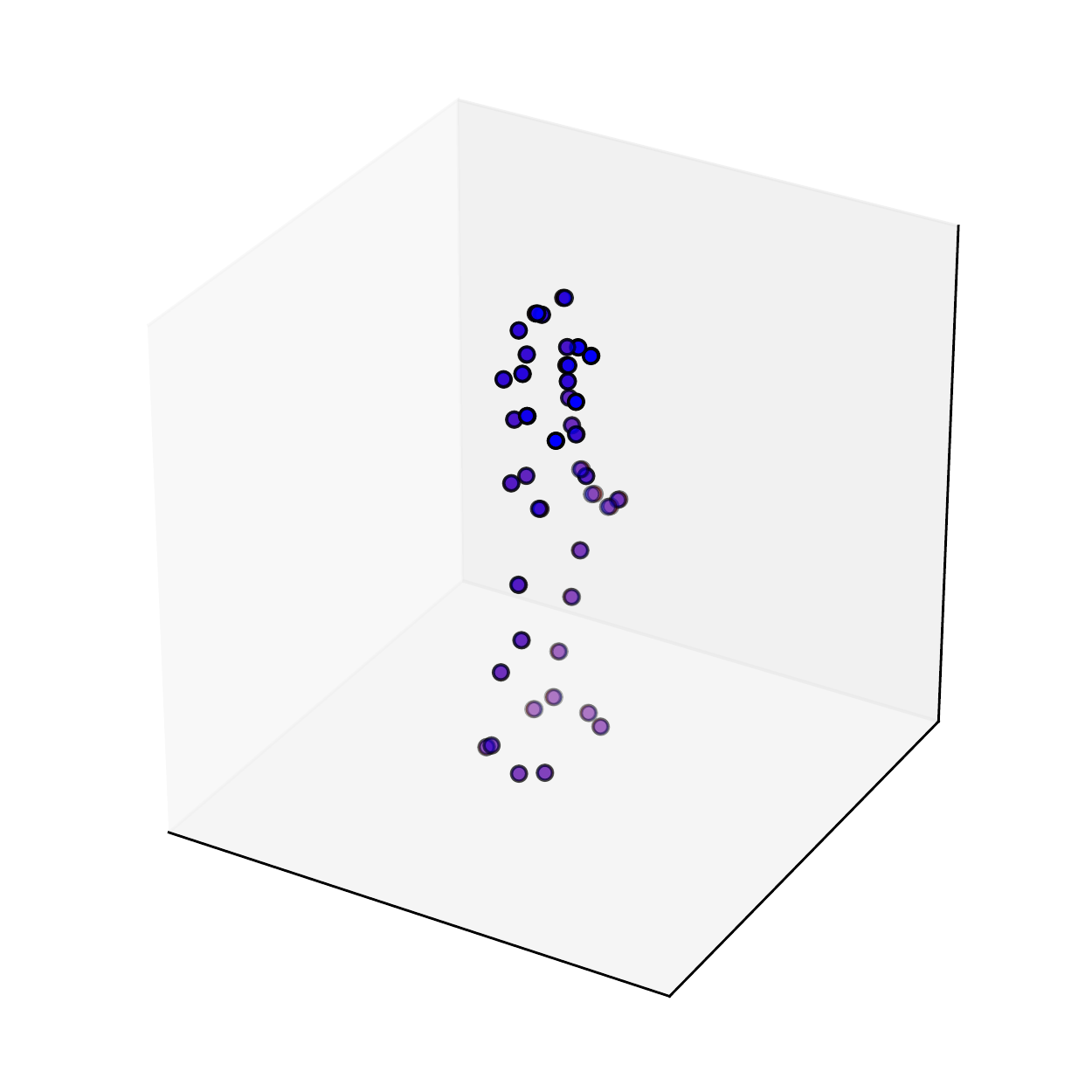} &
    \includegraphics[width=0.13\linewidth]{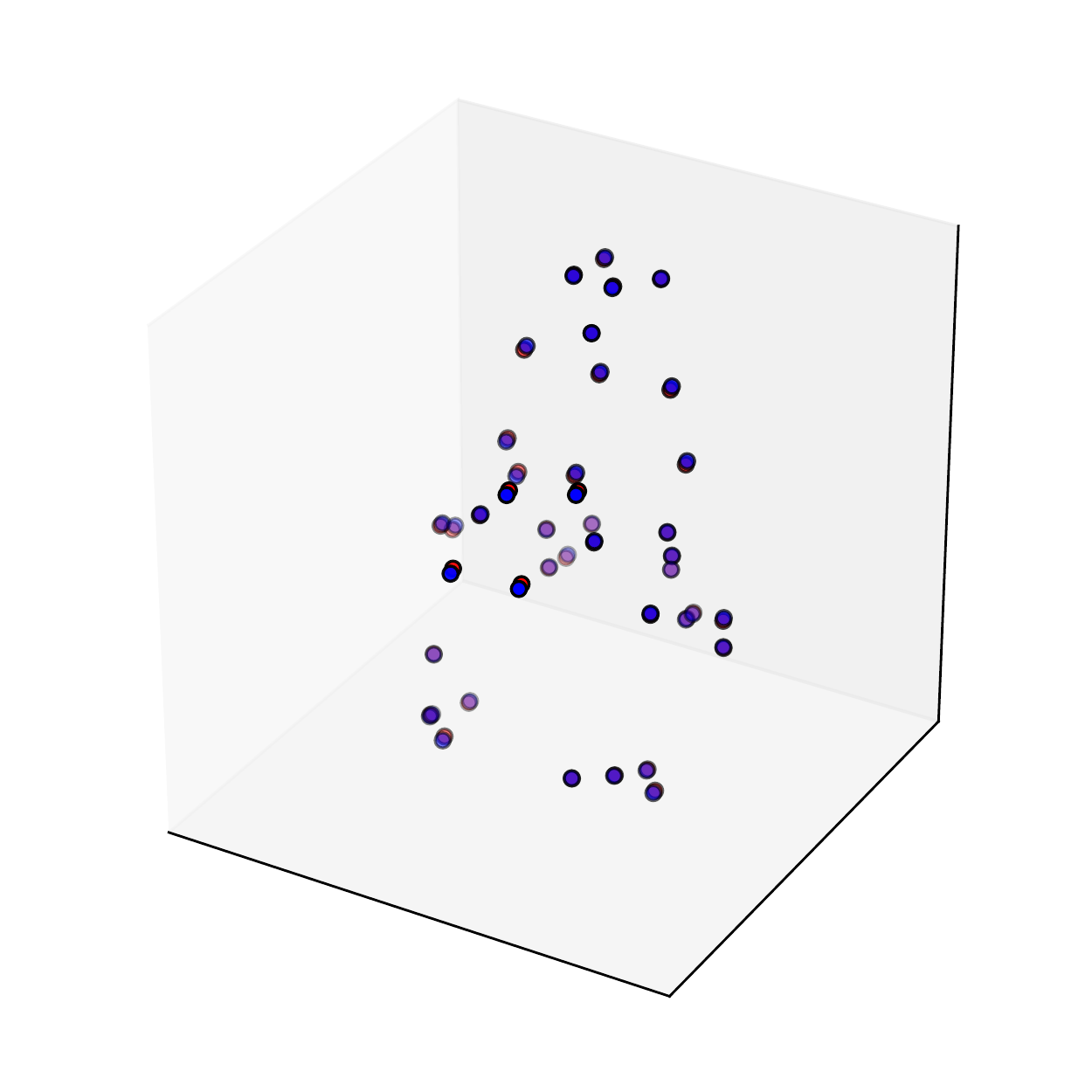} & 
    \includegraphics[width=0.13\linewidth]{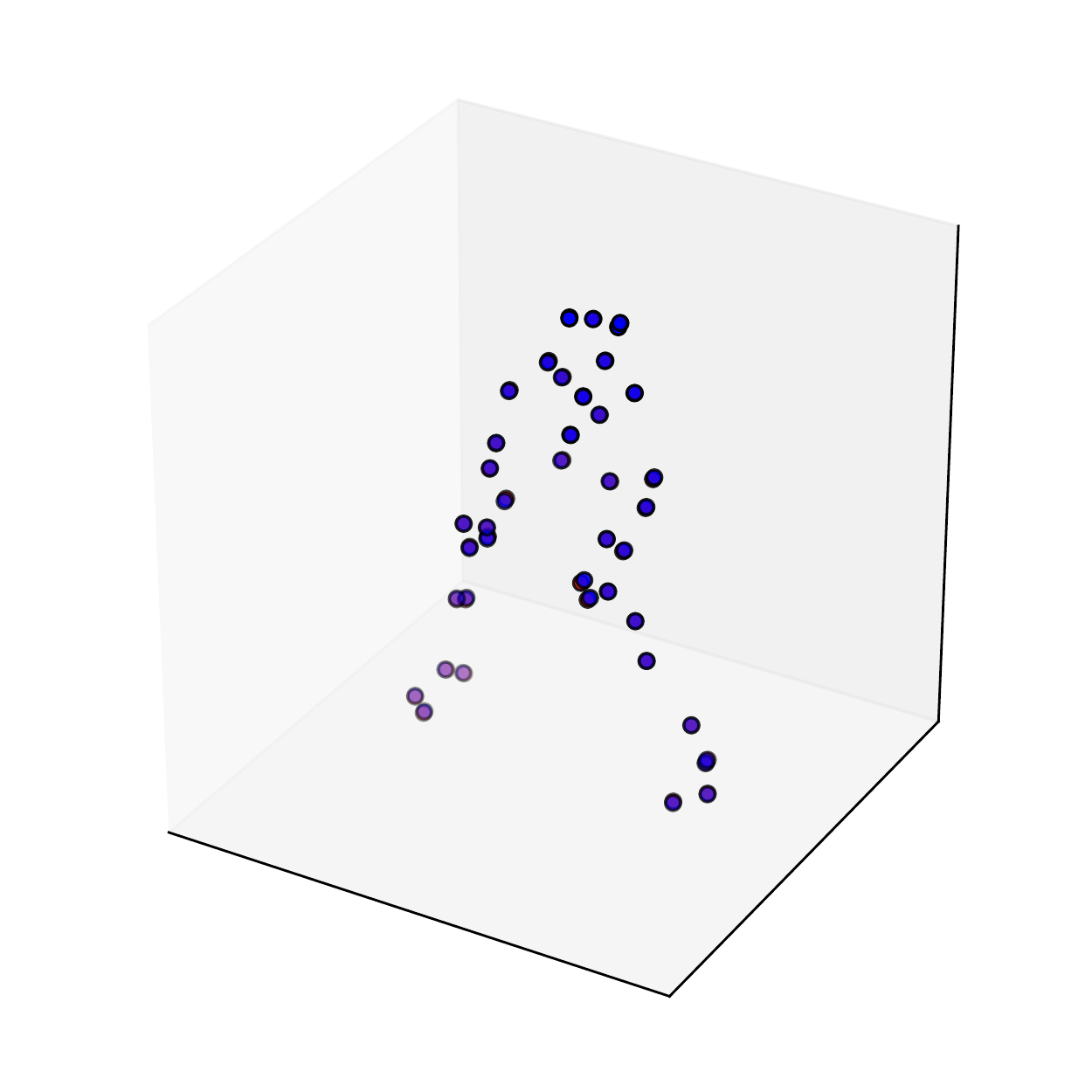} &
    \includegraphics[width=0.13\linewidth]{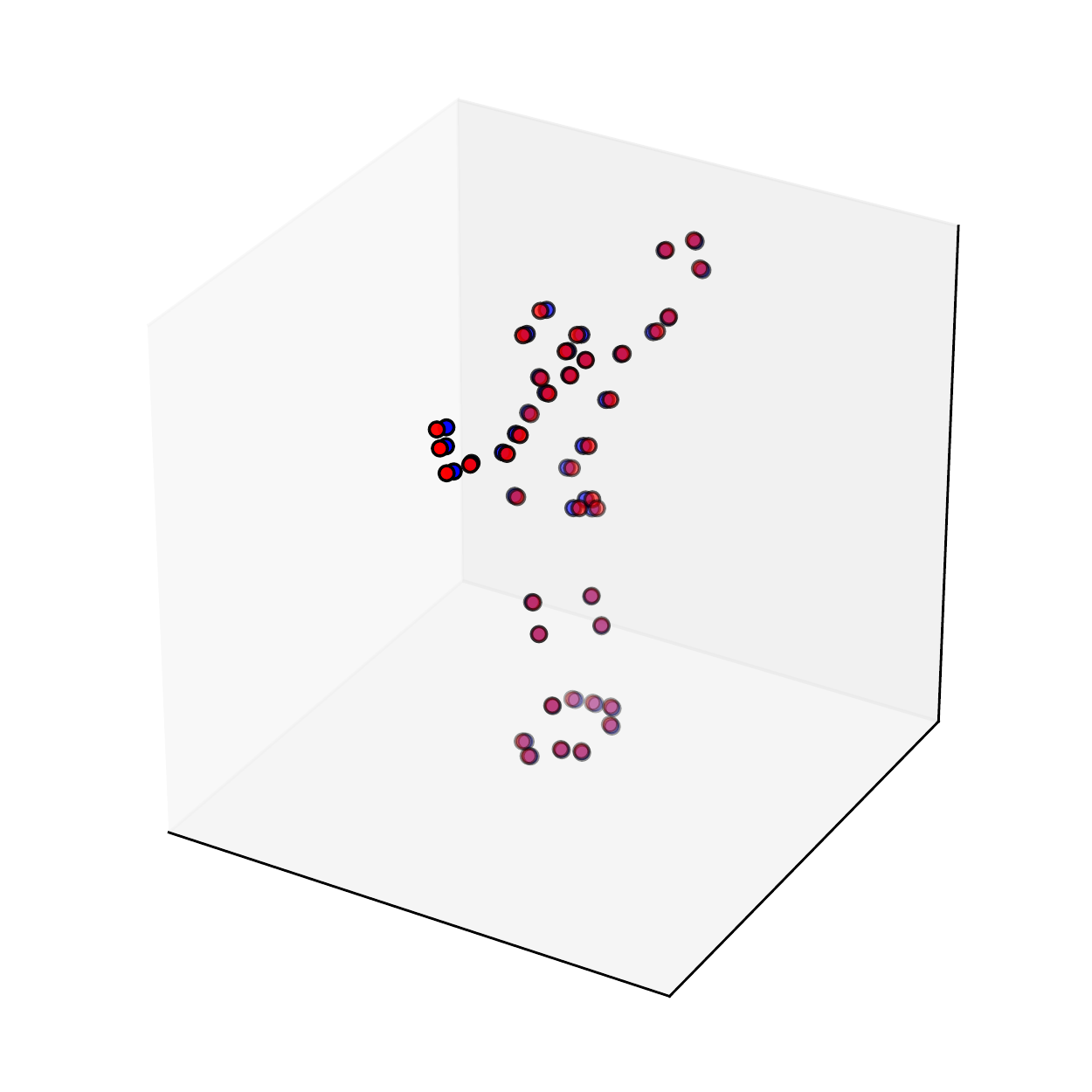} &
    \includegraphics[width=0.13\linewidth]{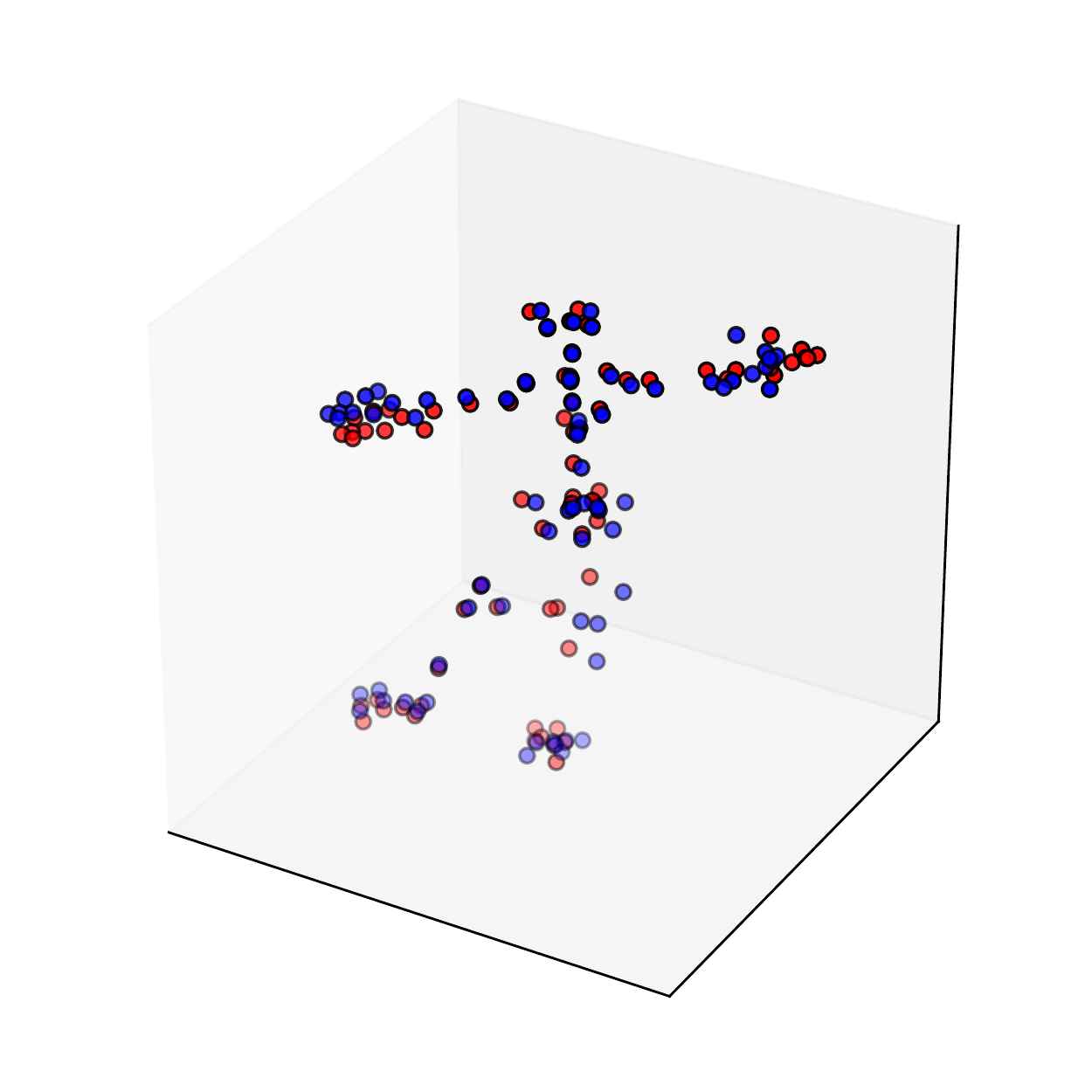} &
    \includegraphics[width=0.13\linewidth]{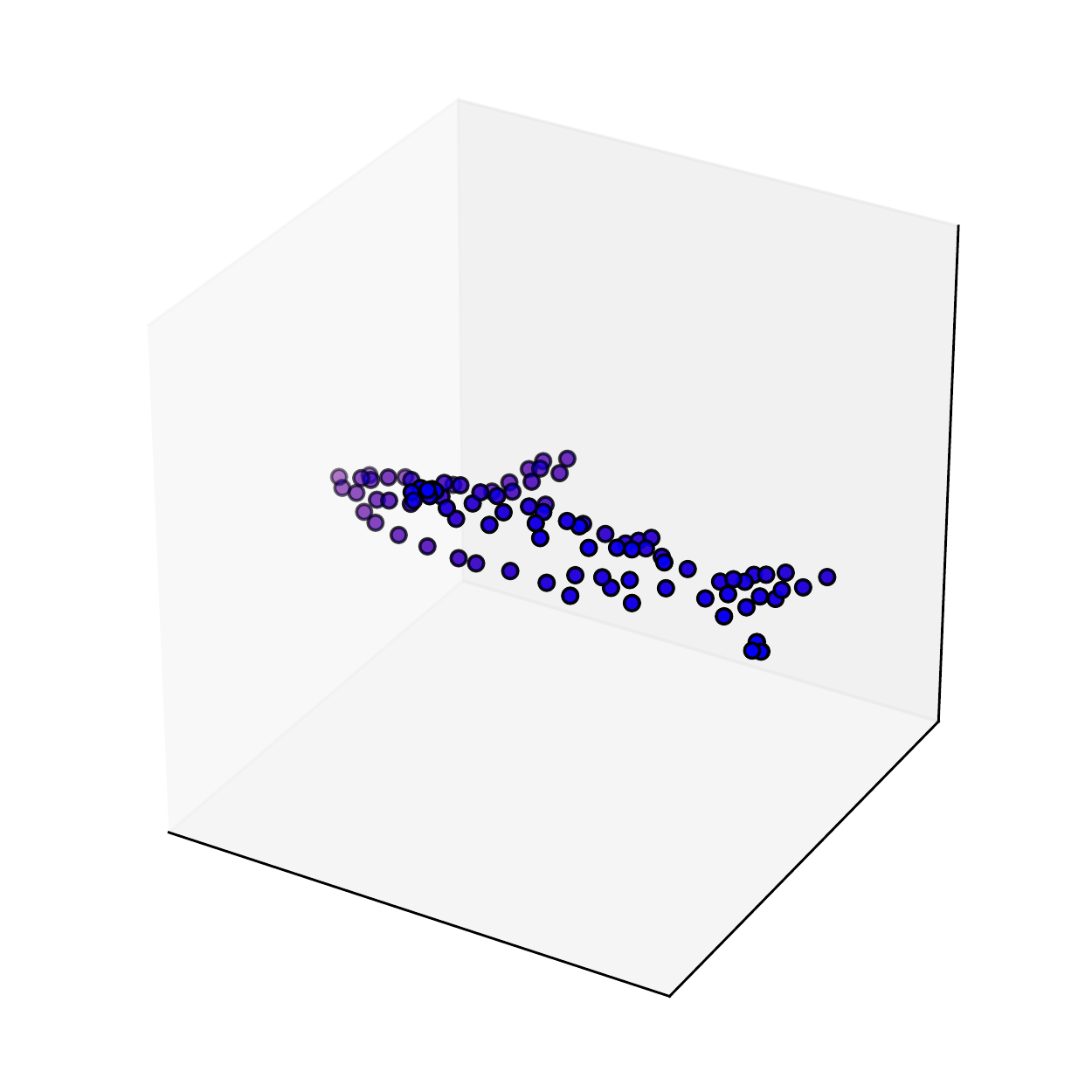}\\
\end{tabular}
    \caption{Results on NRSfM synthetic short sequences. \textcolor{blue}{blue}: points reconstructed by PAUL; \textcolor{red}{red}: groundtruth points. Note the recovered points mostly overlaps with the groundtruth points, indicating the good performance of PAUL.}
    \label{fig:synthetic}
\end{figure*}

\noindent\textbf{auto-encoder v.s. decoder-only lifting.}
In the main paper, we showed that our auto-encoder-based approach (PAUL) empirically outperforms the decoder-only baseline (\ie ADL). The theoretical difference between the two is that auto-encoders additionally enforce the existence of a continuous mapping from the 3D shape to the low-dimensional latent space through learning the encoder network. The implication of this additional constraint is that it encourages shapes with small variation to stay close in the latent space. Consequently it improves the uniqueness of the 2D-3D lifting solution, as it implicitly forces each 3D shape to have a unique latent representation. 

We analyze the difference by visualizing the latent space for both PAUL and ADL. We set the bottleneck dimension as 2 for both methods, and plot the 2D latent code for each sample in the input sequence (see Fig.~\ref{fig:latent_space}). The color (from dark blue to bright yellow) for each point represents its temporal order in the sequence. Since the motion is by nature smooth, temporarily close frames have smaller 3D shape variation, thus ideally their corresponding latent codes should be close to each other. This implies that a continuous code trajectory is expected in the visualization.

In Fig.~\ref{fig:latent_space}\textcolor{red}{(a)}, we first compare PAUL and ADL on short sequences with smooth camera trajectory. As expected, we observe that the reconstructed 2D codes from PAUL indeed forms one or a small number of continuous trajectories. We can also observe recurrent motion from the loop in the code trajectory on the drink sequence. In comparison, without the constraint from the encoder, ADL produces a set of more spread-out latent codes, which forms a number of shorter trajectories in a less interpretable spatial order.  

One may argue that since a 2D-3D encoder is also a continuous mapping, it should partially fullfill the role of an encoder to encourage smoothness of the latent space. However, the counterargument to rely on the inductive-bias from a 2D-3D encoder is -- since 2D projection is a combination of 3D shape and camera pose, small variation in 3D shape does not necessarily leads to small variation in 2D. Hitherto, the continuity of the output of a 2D-3D encoder conditioned on 2D inputs does not translates to the continuity of codes with respect to 3D shapes. To support this counterargument, we conduct experiment on a sequence with random camera motion (see Fig.~\ref{fig:latent_space}\textcolor{red}{(b)}) which means temporal adjacent frames would have very different 2D projections. ADL with a 2D-3D encoder (adopted from Deep NR\SfM~\cite{ck19}) loses the trajectory-like structure in its latent space, which PAUL preserves.

\section*{II. Details for handling missing data}
As shown in Sec. \textcolor{red}{4.3}, we use the extension proposed by Wang~\etal~\cite{wang2020deep} to remove translation and rewrite the projection equation into a bilinear form:
\begin{equation}
\Tilde{\W}\M = \R_\text{xy}\Tilde{\S}\M,
\label{eq:proj_md}
\end{equation}
where $\Tilde{\W}$, $\Tilde{\S}$ denotes the adaptively normalized $\W$, $\S$ according to the visibility mask $\M$. With this new projection equation, we extend the expression of $\mathcal{L}_\text{recon. AE}$ and $\mathcal{L}_\text{recon. AD}$ to account for missing data.

First, due to missing data, the unknown values in $\S_\text{cam}$ are no longer only the depth $\mathbf{z}$, but also includes x-y coordinates of the missed keypoints. Thus we introduce a new free variable $\S_\text{cam}'\in\mathbb{R}^{3\times P}$ to optimize instead of $\mathbf{z}$, and $\Tilde{\S}_\text{cam}$ could be expressed as a fusion between unknown $\S_\text{cam}'$ and known $\W$:
\begin{equation}
    \Tilde{\S}_\text{cam} = \Tilde{\S}_\text{cam}' (1-\M) + \begin{bmatrix} \Tilde{\W} \\ \mathbf{0}^\top \end{bmatrix} \M.
\end{equation}
We note that the above formulation are expressed in adaptive normalized form as in \eqref{eq:proj_md}.

Then, the loss functions are extended to be:
\begin{equation}
\begin{aligned}
    \mathcal{L}_\text{recon. AE} = & \|\widetilde{f_\text{d}\circ f_\text{e} \circ f_\text{d} (\bvarphi)} - \R^\top \Tilde{\S}_\text{cam} \|_F \\
    \mathcal{L}_\text{recon. AD} = & \|\widetilde{f_\text{d} (\bvarphi)} - \R^\top \Tilde{\S}_\text{cam} \|_F
\end{aligned}
\end{equation}
where the operator $\sim$ denotes the same adaptive normalization as defined in (\textcolor{red}{11}).

\section*{III. Additional results}
\begin{figure}
    \centering
    \includegraphics[width=0.85\linewidth]{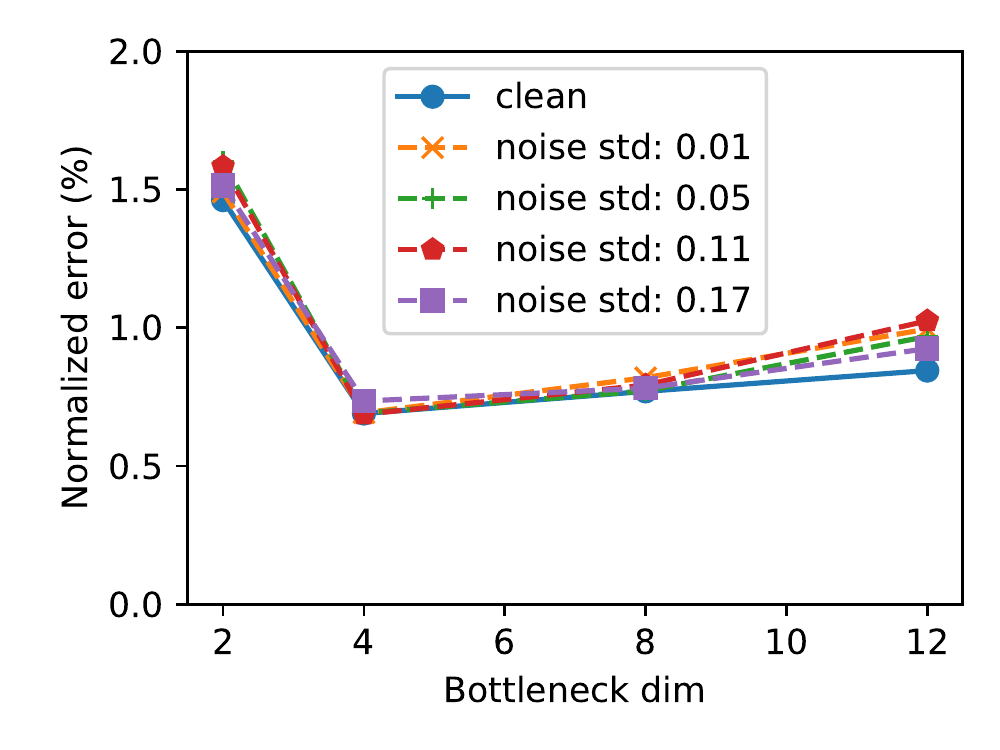}
    \caption{Results on CMU motion capture dataset S70 with different level of noise and bottleneck dimension.}
    \label{fig:noise}
\end{figure}

\noindent \textbf{Robustness to noise.}
We investigate the robustness of PAUL against noise by perturbed the groundtruth 2D keypoints using Gaussian noise with different standard deviation. We further investigate the implication of using different bottleneck dimension when learning from noisy data. As shown in Fig.~\ref{fig:noise}, on CMU motion capture dataset, PAUL learned with clean data produces stable reconstruction accuracy with bottleneck dimension from 4-12. When noise is inserted to the training data, PAUL still keeps similar performance when the bottleneck dimension is relatively small (\eg 4-8). The accuracy decrease becomes more noticeable only when the bottleneck dimension is large (\eg 12). This indicates that PAUL overall is robust against random Gaussian noise, and a good practise to apply PAUL to real data is to start from smaller bottleneck dimension, which provides stronger constraint to denoise the reconstruction.


{\small
\bibliographystyle{ieee_fullname}
\bibliography{egbib}
}

\end{document}